\documentclass[12pt,3p,authoryear]{elsarticle}

\usepackage[table]{xcolor}
\usepackage{hyperref}
\usepackage{url}
\usepackage{amsthm}
\usepackage{amsfonts}
\usepackage{amsmath}
\usepackage{amssymb}
\usepackage{mathtools} 
\usepackage[ruled,vlined]{algorithm2e}
\usepackage{graphicx, subfigure}
\usepackage{booktabs}
\usepackage{multirow}
\usepackage{xcolor}
\definecolor{ForestGreen}{RGB}{0,0,0}

\usepackage{lipsum}
\usepackage{dirtytalk}
\usepackage{caption} 

\newcommand\btheta{\boldsymbol \theta}

\long\def\EDIT#1{{\color{black}{#1}\color{black}}}
\long\def\EDITtwo#1{{\color{ForestGreen}{#1}\color{black}}}

\newcommand{\model}[1]{\texttt{#1}}

\newcommand{\ARIMA}{\model{ARIMA}}
\newcommand{\ARone}{\model{AR1}}
\newcommand{\ARxone}{\model{ARx1}}
\newcommand{\LEAR}{\model{LEAR}}
\newcommand{\LEARx}{\model{LEARx}}
\newcommand{\TBATS}{\model{TBATS}}
\newcommand{\SEATS}{\model{SEATS}}

\newcommand{\DNN}{\model{DNN}}
\newcommand{\NBEATS}{\model{NBEATS}}
\newcommand{\NBEATSx}{\model{NBEATSx}}

\newcommand{\NBEATSi}{\model{NBEATS-I}}
\newcommand{\NBEATSxg}{\model{NBEATSx-G}}
\newcommand{\NBEATSxi}{\model{NBEATSx-I}}

\newcommand{\MQCNN}{\model{MQCNN}}

\newcommand{\ESRNN}{\model{ESRNN}}

\newcommand{\RNN}{\model{RNN}}

\newcommand{\DilRNN}{\model{DilRNN}}

\newcommand{\Identity}{\model{Identity}}
\newcommand{\TCN}{\model{TCN}}
\newcommand{\WaveNet}{\model{WaveNet}}

\newcommand{\FCNN}{\model{FCNN}}

\newcommand{\SeqtoSeq}{\model{Seq2Seq}}

\newcommand{\ADAM}{\model{ADAM}}
\newcommand{\SGD}{\model{SGD}}
\newcommand{\HYPEROPT}{\model{HYPEROPT}}

\newcommand{\SeLU}{\model{SeLU}}
\newcommand{\PreLU}{\model{PreLU}}
\newcommand{\ReLU}{\model{ReLU}}
\newcommand{\TanH}{\model{TanH}}
\newcommand{\LReLU}{\model{LReLU}}
\newcommand{\Sigmoid}{\model{Sigmoid}}
\newcommand{\SoftPlus}{\model{SoftPlus}}

\newcommand{\PyTorch}{\model{PyTorch}}

\newcommand{\dataset}[1]{\texttt{#1}}

\newcommand{\NP}{\dataset{NP}}
\newcommand{\PJM}{\dataset{PJM}}
\newcommand{\BE}{\dataset{EPEX-BE}}
\newcommand{\FR}{\dataset{EPEX-FR}}
\newcommand{\DE}{\dataset{EPEX-DE}}
\newcommand{\EPEX}{\dataset{EPEX}}

\journal{International Journal of Forecasting}

\begin{document}

\begin{frontmatter}

\title{Neural basis expansion analysis with exogenous variables: Forecasting electricity prices with NBEATSx}

\affiliation[inst1]{organization={Auton Lab, School of Computer Science, Carnegie Mellon University},
            city={Pittsburgh},
            state={Pennsylvania},
            country={USA}}
            
\affiliation[inst2]{organization={Department of Operations Research and Business Intelligence, Wroc\l{}aw University of Science and Technology},
            city={Wroc\l{}aw},
            country={Poland}}    

\author[inst1]{Kin G.\ Olivares}
\ead{kdgutier@cs.cmu.edu}
\cortext[cor1]{Corresponding author}
\author[inst1]{Cristian Challu}
\author[inst2]{Grzegorz Marcjasz}
\author[inst2]{Rafa\l{} Weron}
\author[inst1]{Artur Dubrawski}

\begin{abstract}
We extend the \emph{neural basis expansion analysis} (NBEATS) to incorporate exogenous factors. The resulting method, called NBEATSx, improves on a well performing deep learning model, extending its capabilities by including exogenous variables and allowing it to integrate multiple sources of useful information. 
To showcase the utility of the NBEATSx model, we conduct a comprehensive study of its application to electricity price forecasting (EPF) tasks across a broad range of years and markets. 
We observe state-of-the-art performance, significantly improving the forecast accuracy by nearly 20\% over the original NBEATS model, and by up to 5\% over other well established statistical and machine learning methods specialized for these tasks. Additionally, the proposed neural network has an interpretable configuration that can structurally decompose time series, visualizing the relative impact of trend and seasonal components and revealing the modeled processes' interactions with exogenous factors.
To assist related work we made the code available in \EDIT{a dedicated repository}. 
\end{abstract}



\begin{keyword}
Deep Learning \sep NBEATS and NBEATSx models \sep Interpretable neural network \sep Time series decomposition \sep Fourier series \sep Electricity price forecasting
\end{keyword}

\end{frontmatter}


\vspace{10mm}
\section{Introduction}
In the last decade, a significant progress has been made in the application of deep learning to forecasting tasks, with models such as the \emph{exponential smoothing recurrent neural network} (\ESRNN;~\citealt{smyl2020esrnn}) and the \emph{neural basis expansion analysis} (\NBEATS;~\citealt{oreshkin2020nbeats}), outperforming classical statistical approaches in the recent 
M4 competition \citep{makridakis2020m4_competition}. Despite this success we still identify two possible improvements, namely the integration of time-dependent exogenous variables as their inputs and the interpretability of the neural network outputs.

Neural networks have proven powerful and flexible, yet there are several situations where our understanding of the model's predictions can be as crucial as their accuracy, which constitutes a barrier for their wider adoption. The interpretability of the algorithm's outputs is critical because it encourages trust in its predictions, improves our knowledge of the modeled processes, and provides insights that can improve the method itself.

Additionally, the absence of time-dependent covariates makes these powerful models unsuitable for many applications. For instance, Electricity Price Forecasting (EPF) is a task where covariate features are fundamental to obtain accurate predictions. For this reason, we chose this challenging application as a test ground for our proposed forecasting methods.

In this work, we address the two mentioned limitations by first extending the neural basis expansion analysis, allowing it to incorporate temporal and static exogenous variables. And second, by further exploring the interpretable configuration of \NBEATS \ and showing its use as a time-series signal decomposition tool. We refer to the new method as \NBEATSx. The main contributions of this paper include:

\begin{enumerate}[(i)]
	\item \textbf{Incorporation of Exogenous Variables:}
	We propose improvements to the \NBEATS \ model to incorporate time dependent as well as static exogenous variables. For this purpose, we have designed a special substructure built with convolutions, to clean and encode useful information from these covariates, while respecting time dependencies present in the data. These enhancements greatly improve the accuracy of the \NBEATS \ method, and extend its interpretability capabilities, so rare in neural forecasting.
	\item \textbf{Interpretable Time Series Signal Decomposition:}
	Our method combines the power of non-linear transformations provided by neural networks with the flexibility to model multiple seasonalities and simultaneously account for interaction events such as holidays and other covariates, all while remaining interpretable. The extended \NBEATSx \ architecture allows to decompose its predictions into the classic set of level, trend, and seasonality, and identify the effects of exogenous covariates.
	\item \textbf{Time Series Forecasting Comparison:}
	We showcase the use of \NBEATSx \ model on five EPF tasks achieving state-of-the-art performance on all of the considered datasets. 
	We obtain accuracy improvements of almost 20\% in comparison to the original \NBEATS \ and \ESRNN \ architectures, and up to 5\% over other well-established machine learning, EPF-tailored methods~\citep{lago2021epftoolbox}.
\end{enumerate}

The remainder of the paper is structured as follows. Section \ref{section:literature} reviews relevant literature on the developments and applications of deep learning to sequence modeling and current approaches to EPF.
Section \ref{section:model} introduces mathematical notation and describes the \NBEATSx \ model. Section \ref{section:experiments} explores our model's application to time series decomposition and forecasting over a broad range of electricity markets and time periods. Finally, Section \ref{section:conclusion} discusses possible directions for future research, wraps up the results, and concludes the paper.

\newpage
\section{Literature Review} \label{section:literature}
\subsection{Deep Learning and Sequence Modeling}

The Deep Learning methodology (DL) has demonstrated a significant utility in solving sequence modeling problems with applications to natural language processing, audio signal processing, and computer vision. This subsection summarizes the critical DL developments in sequence modeling, that are building blocks of the \NBEATS \ and \ESRNN \ architectures.


For a long time, sequence modeling with neural networks and \emph{Recurrent Neural Networks} (\RNN s; \citealt{elman90rnn})
were treated as synonyms. The hidden internal activations of the RNNs propagated through time provided these models with the ability to encode the observed past of the sequence. This explains their great popularity in building different variants of the \emph{Sequence-to-Sequence} models (\SeqtoSeq) applied to natural language processing \citep{graves2013seq2seq}, and machine translation \citep{sutskever2014seq2seq_translation}. Most progress on RNNs was made possible by architectural innovations and novel training techniques that made their optimization easier.


The adoption of convolutions and skip-connections within the recurrent structures were important precursors for new advancements in sequence modeling, as using deeper representations endowed longer effective memory for the models. Examples of such precursors could be found in \WaveNet \ for audio generation and machine translation \citep{vandenoord2016wavenet}, as well as the \emph{Dilated Recurrent Neural Network} (\DilRNN; \citealt{chang2017dilatedRNN}) and the \emph{Temporal Convolutional Network} (\TCN; \citealt{zico2018tcnn}).


Nowadays, \SeqtoSeq \ models and their derivatives can learn complex nonlinear temporal dependencies efficiently; its use in the time series analysis domain has been a great success. \SeqtoSeq \ models have recently showed better forecasting performance than classical statistical methods, while greatly simplifying the forecasting systems into single-box models, such as the \emph{Multi Quantile Convolutional Neural Network} (\MQCNN; \citealt{wen2017mqrcnn}), the \emph{Exponential Smoothing Recurrent Neural Network} (\ESRNN; \citealt{smyl2020esrnn}), or the \emph{Neural Basis Expansion Analysis} (\NBEATS; \citealt{oreshkin2020nbeats}). For quite a while, the academia resisted to broadly adopt these new methods \citep{makriadakis2018concerns}, although their evident success in challenges such as the M4 competition has motivated their wider adoption by the forecasting research community \citep{benidis2020dl_timeseries_review2}. 


\subsection{Electricity Price Forecasting}


The Electricity Price Forecasting (EPF) task aims at predicting the spot (balancing, intraday, day-ahead) and forward prices in wholesale markets. Since the workhorse of short-term power trading is the day-ahead market with its conducted once-per-day uniform-price auction \citep{may:tru:18}, the vast majority of research has focused on predicting electricity prices for the 24 hours of the next day, either in a point \citep{weron2014epf_survey1,  lago2021epftoolbox} or a probabilistic setting \citep{nowotarski2018epf_survey2}. There also are studies on EPF for very short-term \citep{nar:zie:20}, as well as mid- and long-term horizons \citep{zie:ste:18}. The recent expansion of renewable energy generation and large-scale battery storage has induced complex dynamics to the already volatile electricity spot prices, turning the field into a prolific subject on which to test novel forecasting ideas and trading strategies \citep{chi:etal:18, gianfreda2020epf_res, uniejewski2021quantile_reg}. 




Out of the numerous approaches to EPF developed over the last two decades,  
two classes of models are of particular importance when predicting day-ahead prices -- statistical (also called econometric or technical analysis), in most cases based on linear regression, and computational intelligence (also referred to as artificial intelligence, non-linear or machine learning), with neural networks being the fundamental building block. Among the latter, many of the recently proposed methods utilize deep learning (\citealt{Wang2016b, lago2018DNN,Marcjasz2020}), or are hybrid solutions, that typically comprise data decomposition, feature selection, clustering, forecast averaging and/or heuristic optimization to estimate the model (hyper)parameters \citep{Nazar2018,LiBecker2021}. 

Unfortunately, as argued by \cite{lago2021epftoolbox}, the majority of the neural network EPF related research suffers from too short and limited to a single market test periods, lack of well performing and established benchmark methods, and/or incomplete descriptions of the pipeline and training methodology resulting in poor reproducibility of the results.
To address these shortcomings, our models are compared across two-year out-of-sample periods from five power markets and using two highly competitive benchmarks recommended in previous studies: the \emph{Lasso Estimated Auto-Regressive} (\LEAR) model and a (relatively) parsimonious \emph{Deep Neural Network} (\DNN).


\section{NBEATSx Model} \label{section:model}
\begin{figure}[tb]
\centering
\includegraphics[width=1.0\linewidth]{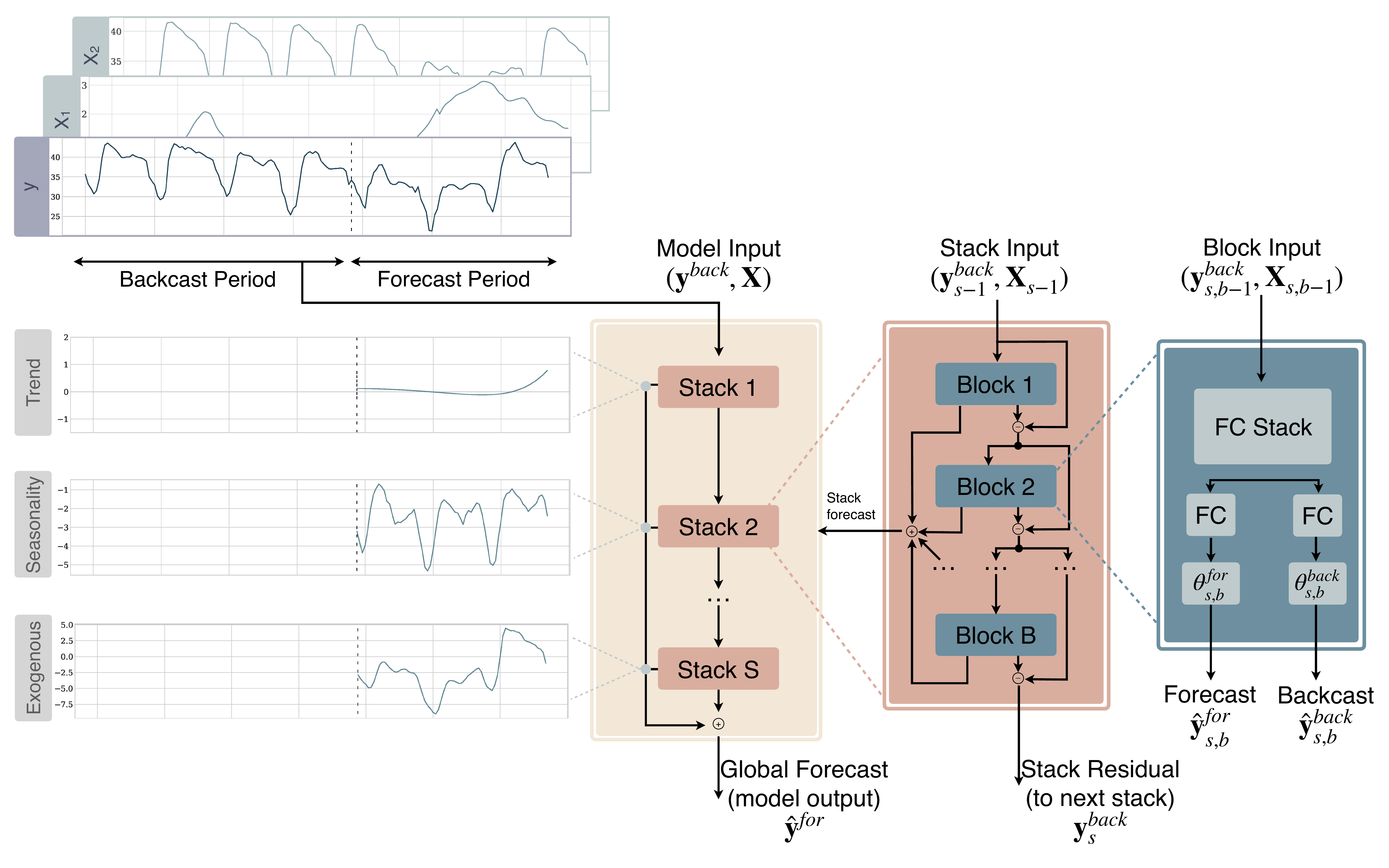}
\caption{Building blocks of the \NBEATSx\ are structured as a system of multilayer fully connected networks with ReLU based nonlinearities. 
Blocks overlap using the doubly residual stacking principle for the backcast \EDIT{$\mathbf{\hat{y}}^{back}_{s,b}$ and forecast $\mathbf{\hat{y}}^{for}_{s,b}$ outputs of the $b$-th block within the $s$-th stack. The final predictions $\mathbf{\hat{y}}^{for}$ are composed by aggregating the outputs of the stacks.} 
} \label{fig:nbeatsx_architecture} 
\end{figure}

As a general overview, the \NBEATSx\ framework decomposes the objective signal by performing separate local nonlinear projections of the target data onto basis functions across its different blocks. Figure~\ref{fig:nbeatsx_architecture} depicts the general architecture of the model. Each block consists of a \emph{Fully Connected Neural Network} (\FCNN; \citealt{rosenblatt1961principles}) which learns expansion coefficients for the backcast and forecast elements. The backcast model is used to clean the inputs of subsequent blocks, while the forecasts are summed to compose the final prediction. The blocks are grouped in stacks. Each of the potentially multiple stacks specializes in a different variant of basis functions.

To continue the description of the \NBEATSx, we introduce the following notation: the objective signal is represented by the vector $\mathbf{y}$, the inputs for the model are the backcast window vector $\mathbf{y}^{back}$ of length $L$, and the forecast window vector $\mathbf{y}^{for}$ of length $H$; where $L$ denotes the length of the lags available as classic autoregressive features, and $H$ is the forecast horizon treated as the objective. While the original \NBEATS\ only admits as regressor the backcast period of the target variable $\mathbf{y}^{back}$, the \NBEATSx\ incorporates covariates in its analysis denoted with the matrix $\mathbf{X}$. Figure~\ref{fig:nbeatsx_architecture} shows an example where the target variable is the hourly electricity price,  the backcast vector has a length $L$ of 96 hours, and the forecast horizon $H$ is 72 hours, in the example, the covariate matrix $\mathbf{X}$ is composed of wind power production and electricity load. For the EPF comparative analysis of Section~\ref{subsection:forecasting_results} the horizon considered is $H=24$ that corresponds to day-ahead predictions, while backcast inputs $L=168$ correspond to a week of lagged values.

\EDIT{For its predictions, the \NBEATS\ model only receives a local vector of inputs corresponding to the backcast period, making the computations exceptionally fast. The model can still represent longer time dependencies through its local inputs from the exogenous variables; for example, it can learn long seasonal effects from calendar variables.}

\EDIT{Overall, as shown in Figure~\ref{fig:nbeatsx_architecture}, the \NBEATSx\ is composed of $S$ stacks of $B$ blocks each, the input $\mathbf{y}^{back}$ of the first block consists of $L$ lags of the target time series $\mathbf{y}$ and the exogenous matrix $\mathbf{X}$, while the inputs of each of the subsequent blocks include residual connections with the backcast output of the previous block. We will describe in detail in the next subsections the blocks, stacks, and model predictions.} 

\clearpage
\subsection{Blocks}

For a given $s$-th stack and $b$-th block within it, the \NBEATSx\ model performs two transformations, depicted in the blue rectangle of Figure~\ref{fig:nbeatsx_architecture}. The first transformation, defined in Equation~(\ref{equation:block_transformations}), takes the input data $ (\mathbf{y}^{back}_{s,b-1}, \mathbf{X}_{s,b-1})$, and applies a \emph{Fully Connected Neural Network} (\FCNN; \citealt{rosenblatt1961principles}) to learn hidden units $\color{ForestGreen} \mathbf{h}_{s,b} \in \mathbb{R}^{N_{h}}$ that are linearly adapted into the forecast $\btheta^{for}_{s,b} \in \mathbb{R}^{N_s}$ and backcast $\btheta^{back}_{s,b}\in \mathbb{R}^{N_s}$ expansion coefficients, with $N_s$ the dimension of the stack basis.
\begin{equation}
\begin{aligned}
    \mathbf{h}_{s,b} &= \mathbf{FCNN}_{s,b}\left(\mathbf{y}^{back}_{s,b-1}, \mathbf{X}_{b-1}\right) \\
    \btheta^{back}_{s,b} &= \textbf{LINEAR}^{back}\left(\mathbf{h}_{s,b}\right) \qquad
    \btheta^{for}_{s,b} &= \textbf{LINEAR}^{for}\left(\mathbf{h}_{s,b}\right)
    \label{equation:block_transformations}
\end{aligned}
\end{equation}
The second transformation, defined in Equation~(\ref{equation:nonlinear_projections}), consists of a basis expansion operation between the learnt coefficients and the block's basis vectors $\mathbf{V}^{back}_{s,b} \in \mathbb{R}^{L \times N_s}$ and $ \mathbf{V}^{for}_{s,b} \in \mathbb{R}^{H \times N_s}$, this transformation results in the backcast $ \mathbf{\hat{y}}^{back}_{s,b}$ and forecast $ \mathbf{\hat{y}}^{for}_{s,b}$ components.
\begin{equation}
    \mathbf{\hat{y}}^{back}_{s,b} 
    = \mathbf{V}^{back}_{s,b} \, \btheta^{back}_{s,b}
    \quad \text{and} \quad
    \mathbf{\hat{y}}^{for}_{s,b} 
    = \mathbf{V}^{for}_{s,b} \,\btheta^{for}_{s,b}
    \label{equation:nonlinear_projections}
\end{equation}

\subsection{Stacks and Residual Connections}

\EDIT{The blocks are organized into stacks using the doubly residual stacking principle, which is described in Equation~(\ref{equation:stacks_connections}) and depicted in the brown rectangle of Figure~\ref{fig:nbeatsx_architecture}.} The residual backcast $ \mathbf{y}^{back}_{s,b+1}$ allows the model to subtract the component associated to the basis of the $s,b$-th stack and block $ \mathbf{V}^{back}_{s,b}$ from $\mathbf{y}^{back}$, which can be also thought of as a sequential decomposition of the modeled signal. In turn, this methodology helps with the optimization procedure as it prepares the inputs of the subsequent layer making the downstream forecast easier. The stack forecast $\mathbf{y}^{for}_{s}$ aggregates the partial forecasts from each block.

\begin{equation}
    \mathbf{y}^{back}_{s,b+1} = \mathbf{y}^{back}_{s,b}-\hat{\mathbf{y}}^{back}_{s,b} 
    \qquad \text{and} \qquad
    \mathbf{\hat{y}}^{for}_{s}
    = \sum^{B}_{b=1} \hat{\mathbf{y}}^{for}_{s,b}
    \label{equation:stacks_connections}
\end{equation}

\subsection{Model predictions}

The final predictions $\hat{\mathbf{y}}^{for}$ of the model, shown in the yellow rectangle of Figure~\ref{fig:nbeatsx_architecture}, are obtained by summation of all the stack predictions.
\begin{equation}
    \mathbf{\hat{y}}^{for} 
    = \sum^{S}_{s=1} \hat{\mathbf{y}}^{for}_{s} 
    \label{equation:predictions}
\end{equation}

The additive generation of the forecast implies a very intuitive decomposition of the prediction components when the bases within the blocks are interpretable.

\clearpage
\subsection{NBEATSx Configurations}
\label{subsection:nbeatsx_configurations}

The original neural basis expansion analysis method proposed two configurations based on the assumptions encoded in the learning algorithm by selecting the basis vectors $\mathbf{V}^{back}_{s,b}$ and $\mathbf{V}^{for}_{s,b}$ used in the blocks from Equation~(\ref{equation:nonlinear_projections}). A mindful selection of restrictions to the basis allows the model to output an interpretable decomposition of the forecasts, \EDIT{while allowing the basis to be freely determined can produce more flexible forecasts by effectively removing any constraints on the form of the basis functions.} 

In this subsection, we present both interpretable and generic configurations, explaining in particular how we propose to include the covariates in each case. \EDIT{We limit ourselves to the analysis of the forecast basis, as the backcast basis analysis is almost identical, only differing by its extension over time. We show an example in \ref{appendix:basis}}. 

\subsubsection{Interpretable Configuration}
\label{subsubsection:nbeatsx_configurations}

The choice of basis vectors relies on time series decomposition techniques that are often used to understand the structure of a given time series and patterns of its variation. 
Work in this area ranges from classical smoothing methods 
and their extensions such as X-11-\ARIMA, 
X-12-\ARIMA, 
and X-13-\ARIMA-\SEATS, 
to modern approaches such as \TBATS~\citep{hyndman2011tbats}.
To encourage interpretability, the blocks within each stack may use harmonic functions, polynomial trends, and exogenous variables directly to perform their projections. 
The partial forecasts of the interpretable configuration are described by Equations~(\ref{equation:interpretable_trend})-(\ref{equation:interpretable_exogenous}).

\begin{equation}
    \hat{\mathbf{y}}^{trend}_{s,b} 
    = \sum_{i=0}^{\color{ForestGreen}{N_{pol}}} \mathbf{t}^{i}  \; \theta^{trend}_{s,b,i}
    \equiv \mathbf{T} \; \btheta^{trend}_{s,b}
\label{equation:interpretable_trend}
\end{equation}

\begin{equation}
    \hat{\mathbf{y}}^{seas}_{s,b} 
    = \sum^{\lfloor H/2 -1 \rfloor}_{i=0} 
    \cos\left(2\pi i \color{ForestGreen}{\frac{\mathbf{t}}{N_{hr}}} \right) \theta^{seas}_{s,b,i} 
    + \sin \left( 2\pi i \color{ForestGreen}{\frac{\mathbf{t}}{N_{hr}}} \right) \theta^{seas}_{s,b,i+\lfloor H/2 \rfloor} 
    \equiv  \mathbf{S} \; \btheta^{seas}_{s,b}
\label{equation:interpretable_seasonality}
\end{equation}

\begin{equation}
    \hat{\mathbf{y}}^{exog}_{s,b} 
    = \sum^{N_x}_{i=0} \mathbf{X}_{i} \, \theta^{exog}_{s,b,i}
    \equiv \mathbf{X} \; \btheta^{exog}_{s,b}  
\label{equation:interpretable_exogenous}
\end{equation}
where the time vector $\mathbf{t}^{\intercal}=[0,1,2,\dots,H-2, H-1]/H$ is defined discretely.  When the basis $\mathbf{V}^{for}_{s,b}$ is $\color{ForestGreen} \mathbf{T}=[\mathbf{1},\mathbf{t},\dots,\mathbf{t}^{N_{pol}}] \in \mathbb{R}^{H \times (N_{pol}+1)}$, where $\color{ForestGreen} N_{pol}$ is the maximum polynomial degree, the coefficients are those of a trend polynomial model. When the bases $\mathbf{V}^{for}_{s,b}$ are harmonic $\color{ForestGreen} \mathbf{S}=[\mathbf{1},\cos(2\pi \frac{\mathbf{t}}{N_{hr}}),\dots, \cos(2 \pi \lfloor H/2 -1  \rfloor \frac{\mathbf{t}}{N_{hr}}), \sin(2\pi \frac{\mathbf{t}}{N_{hr}}),\dots,\sin(2 \pi \lfloor H/2 -1 \rfloor \frac{\mathbf{t}}{N_{hr}})] \in \mathbb{R}^{H \times (H-1)} $, the coefficients vector $ \btheta^{for}_{s,b}$ can be interpreted as Fourier transform coefficients, \EDITtwo{the hyperparameter $N_{hr}$ controls the harmonic oscillations}. The exogenous basis expansion can be thought as a time-varying local regression when the basis is the matrix $ \mathbf{X}=[\mathbf{X}_{1},\dots,\mathbf{X}_{N_x}] \in \mathbb{R}^{H \times N_x} $, where $N_{x}$ is the number of exogenous variables. The resulting models can flexibly reflect common structural assumptions, in particular using the interpretable bases, as well as their combinations. 

In this paper, we propose including one more type of stack to specifically represent exogenous variable basis as described in Equation~(\ref{equation:interpretable_exogenous}) and depicted in Figure~\ref{fig:nbeatsx_architecture}. 
In the original \NBEATS\ framework~(\cite{oreshkin2020nbeats}), the interpretable configuration usually consists of a trend stack followed by a seasonality stack, each containing three blocks. 
Our \NBEATSx\ extension of this configuration consists of three stacks, one of each type of factors (trend, seasonal, exogenous). 
\EDIT{We refer to this interpretable and its enhanced interpretable configuration as the \NBEATSi\ and \NBEATSxi\ models, respectively.} 

\subsubsection{Generic Configuration}

For the generic configuration, the basis of the non linear projection in Equation~(\ref{equation:nonlinear_projections}) corresponds to canonical vectors, that is $\mathbf{V}^{for}_{s,b} = I_{H \times H}$, \EDIT{an identity matrix of dimensionality equal to the forecast horizon $H$ that matches the coefficient's cardinality} $ |\btheta^{for}_{s,b}| = H$. 
\begin{equation}
\begin{aligned}
    \mathbf{\hat{y}}^{gen}_{s,b} 
    = \mathbf{V}^{for}_{s,b} \; \btheta^{for}_{s,b} 
    = \btheta^{for}_{s,b}
    \label{equation:generic_basis}
\end{aligned}
\end{equation}

This basis enables \NBEATSx\ to effectively behave like a classic \emph{Fully Connected Neural Network} (\FCNN).
The output layer of the \FCNN\ inside each block has $H$ neurons, that correspond to the forecast horizon, each producing the forecast for one particular time point of the forecast period. This can be understood as the basis vectors being learned during optimization, allowing the waveform of the basis of each stack to be freely determined in a data-driven fashion. Compared to the interpretable counterpart described in Section~\ref{subsubsection:nbeatsx_configurations}, the constraints on the form of the basis functions are removed. This affords the generic variant more flexibility and power at representing complex data, but it can also lead to less interpretable outcomes and potentially escalated risk of overfitting.

For the \NBEATSx\ model with the generic configuration, we propose a new type of exogenous block that learns a context vector $\mathbf{C}_{s,b}$ from the time-dependent covariates with an \textit{encoder} convolutional sub-structure:
\begin{equation}
\begin{aligned}
    \hat{\mathbf{y}}^{exog}_{s,b} 
    = \sum^{N_{c}}_{i=1} C_{s,b,i} \theta^{for}_{s,b,i} 
    \equiv \mathbf{C}_{s,b} \btheta^{for}_{s,b}
    \qquad \text{with} \qquad
    \mathbf{C}_{s,b} = \text{TCN}(\mathbf{X}) \\
    \label{equation:exogenous_encoder}
\end{aligned}
\end{equation}
In the previous equation, a \emph{Temporal Convolutional Network} (\TCN;~\citealt{zico2018tcnn}) is employed as an \emph{encoder}, but any neural network with a sequential structure will be compatible with the backcast and forecast branches of the model, and could be used as an \emph{encoder}. For example, the \WaveNet~\citep{vandenoord2016wavenet} can be an effective alternative to RNNs as it is also able to capture long term dependencies and interactions of covariates by stacking multiple layers, while dilations help it keep the models computationally tractable. In addition, convolutions have a very convenient interpretation as a weighted moving average signal filters. The final linear projection and the additive composition of the predictions can be interpreted as a \emph{decoder}.

The original \NBEATS\ configuration includes only one generic stack with dozens of blocks, while our proposed model includes both the generic and exogenous stacks, with the order determined via data-driven hyperparameter tuning. We refer to this configuration as the \NBEATSxg\ model.

\subsubsection{Exogenous Variables}



We distinguish the exogenous variables by whether they reflect static or time-dependent aspects of the modeled data.
\EDIT{The \emph{static} exogenous variables carry time-invariant information. When the model is built with common parameters to forecast multiple time series, these variables allow sharing information within groups of time series with similar static variable levels. Examples of static variables include designators such as identifiers of regions, groups of products, among others. 

As for the \emph{time-dependent} exogenous covariates, we discern two subtypes. First, we consider seasonal covariates from the natural frequencies in the data. These variables are useful for \NBEATSx\ to identify seasonal patterns and special events inside and outside the window lookback periods. Examples of these are the trends and harmonic functions from Equation~(\ref{equation:interpretable_trend}) and Equation~(\ref{equation:interpretable_seasonality}). Second, we identify domain-specific temporal covariates unique to each problem. The EPF setting typically includes day-ahead forecasts of electricity load and production levels from renewable energy sources.}

\section{Empirical Evaluation} \label{section:experiments}
\subsection{Electricity Price Forecasting Datasets}

To evaluate our method's forecasting capabilities, we consider short-term electricity price forecasting tasks, where the objective is to predict day-ahead prices. Five major power markets\footnote{For the sake of reproducibility we only consider datasets that are openly accessible in the EPFtoolbox library \url{https://github.com/jeslago/epftoolbox} \citep{lago2021epftoolbox}.} are used in the empirical evaluation, all comprised of hourly observations of the prices and two influential temporal exogenous variables that extend for 2,184 days (312 weeks, six years). From the six years of available data for each market, we hold two years out, to test the forecasting performance of the algorithms. The length and diversity of the test sets allow us to obtain accurate and highly comprehensive measurements of the robustness and the generalization capabilities of the models.

Table~\ref{tab:interpretability:datasets} summarizes the key characteristics of each market. The Nord Pool electricity market (\NP), which corresponds to the Nordic countries exchange, contains the hourly prices and day-ahead forecasts of load and wind generation. The second dataset is the Pennsylvania-New Jersey-Maryland market in the United States (\PJM), which contains hourly zonal prices in the Commonwealth Edison (COMED) and two day-ahead forecasts of load at the system and COMED zonal levels. The remaining three markets are obtained from the integrated European Power Exchange (\EPEX). Belgium (\BE) and France (\FR) markets share the day-ahead forecast generation in France as covariates since it is known to be one of the best predictors for Belgian prices \citep{lago2018market_integration}. 
Finally, the German market (\DE) contains the hourly prices, day-ahead load forecasts, and the country level wind and solar generation day-ahead forecast.

\begin{table}[tb]
\caption{Datasets used in our empirical study. For the five day-ahead electricity markets considered, we report the test period dates and two influential covariate variables.}
\label{tab:interpretability:datasets}
\centering
\footnotesize
    \begin{tabular}[t]{llll} 
    \toprule
    \textsc{Market} &  \textsc{Exogenous Variable 1} & \textsc{Exogenous Variable 2} & \textsc{Test Period}\\
    \midrule
    \NP   & day-ahead load & day-ahead wind generation & 27-12-2016 to 24-12-2018 \\
    \PJM  & 2 day-ahead system load & 2 day-ahead COMED load & 27-12-2016 to 24-12-2018 \\
    \FR   & day-ahead load & day-ahead total France generation & 04-01-2015 to 31-12-2016 \\
    \BE   & day-ahead load & day-ahead total France generation & 04-01-2015 to 31-12-2016 \\
    \DE   & day-ahead zonal load & day-ahead wind and solar generation & 04-01-2016 to 31-12-2017 \\
    \bottomrule
    \end{tabular}
\end{table}%

\begin{figure}[p]
\centering
\includegraphics[width=.85\linewidth]{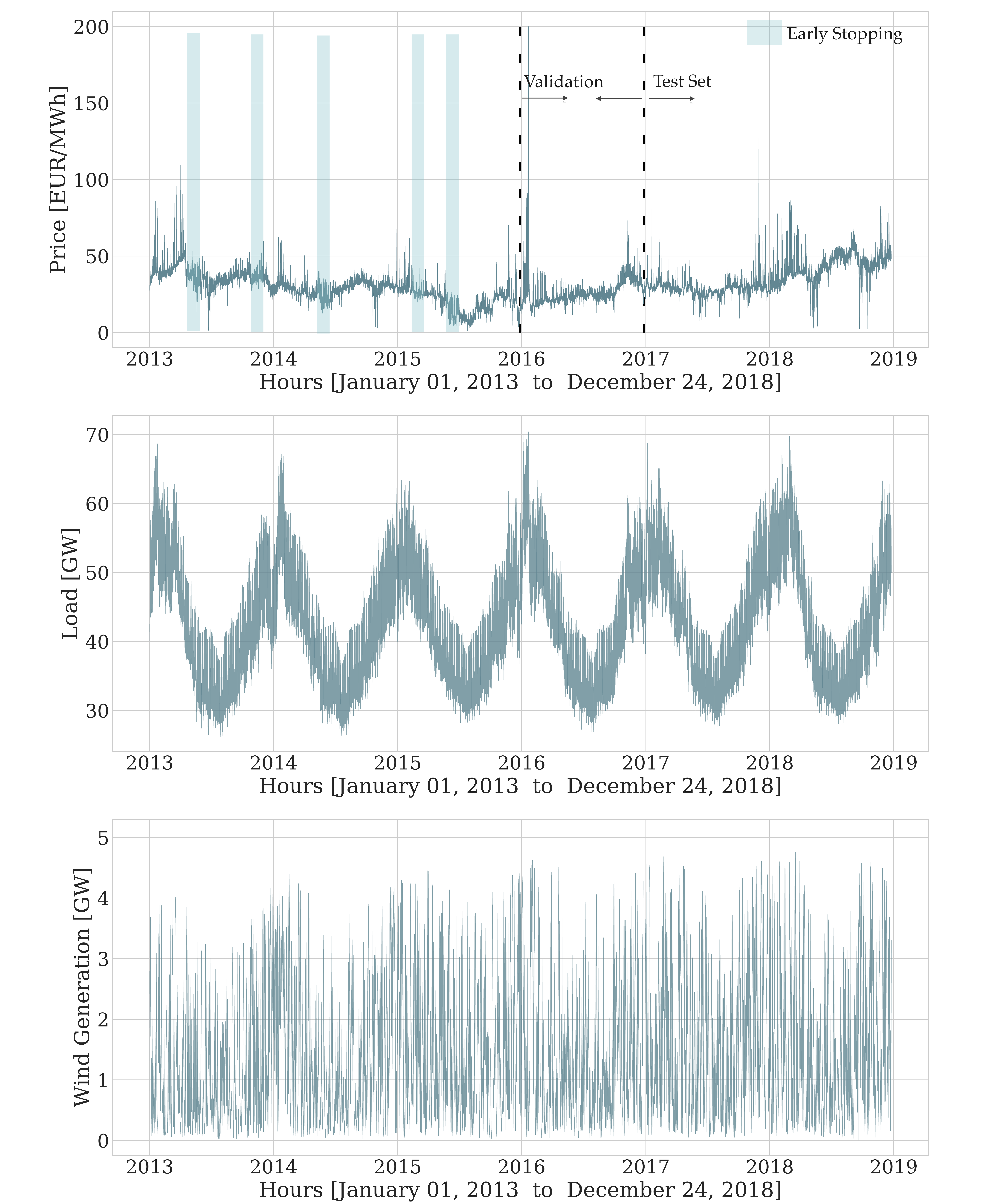}
\caption{The top panel shows the \textit{day-ahead} electricity price time series for the NordPool (\NP) market. The second and third panels show the day-ahead forecast for the system load and wind generation.
The training data is composed of the first four years of each dataset. The validation set is the year that follows the training data (between the first and second dotted lines).  For the held-out test set, the last two years of each dataset are used (marked by the second dotted line). During \emph{evaluation}, we recalibrate the model updating the training set to incorporate all available data before each daily prediction. The recalibration uses an early stopping set of 42 weeks randomly chosen from the updated training set (a sample selection is marked with blue rectangles in the top panel).} 
\label{fig:NP_electricity_datasets}
\end{figure}

Figure \ref{fig:NP_electricity_datasets} displays the \NP\ electricity price time series and its corresponding covariate variables to illustrate the datasets. The \NP\ market is the least volatile among the considered markets, since most of its power comes from hydroelectric generation, renewable source volatility is negligible, and zero spikes are rare. The \PJM\ market is transitioning from coal generation to natural gas and some renewable sources, zero spikes are rare, but the system exhibits higher volatility than \NP. In \BE\ and \FR\ markets, negative prices and spikes are more frequent, and as time passes, these markets begin to show increasing signs of integration. Finally, the \DE\ market shows few price spikes, but the most frequent negative and zero price events, due in great part to the impact of renewable sources.

The exogenous covariates are normalized following best practices drawn from the EPF literature  \citep{weron2018variance_stabilizing}. Preprocessing the inputs of neural networks is essential to accelerate and stabilize the optimization \citep{LeCun1998backprop_tricks}. 

\subsection{Interpretable Time Series Signal Decomposition}

\begin{figure}[tbp]
\centering
\subfigure[NBEATS]{\label{fig:a}
\includegraphics[width=72mm]{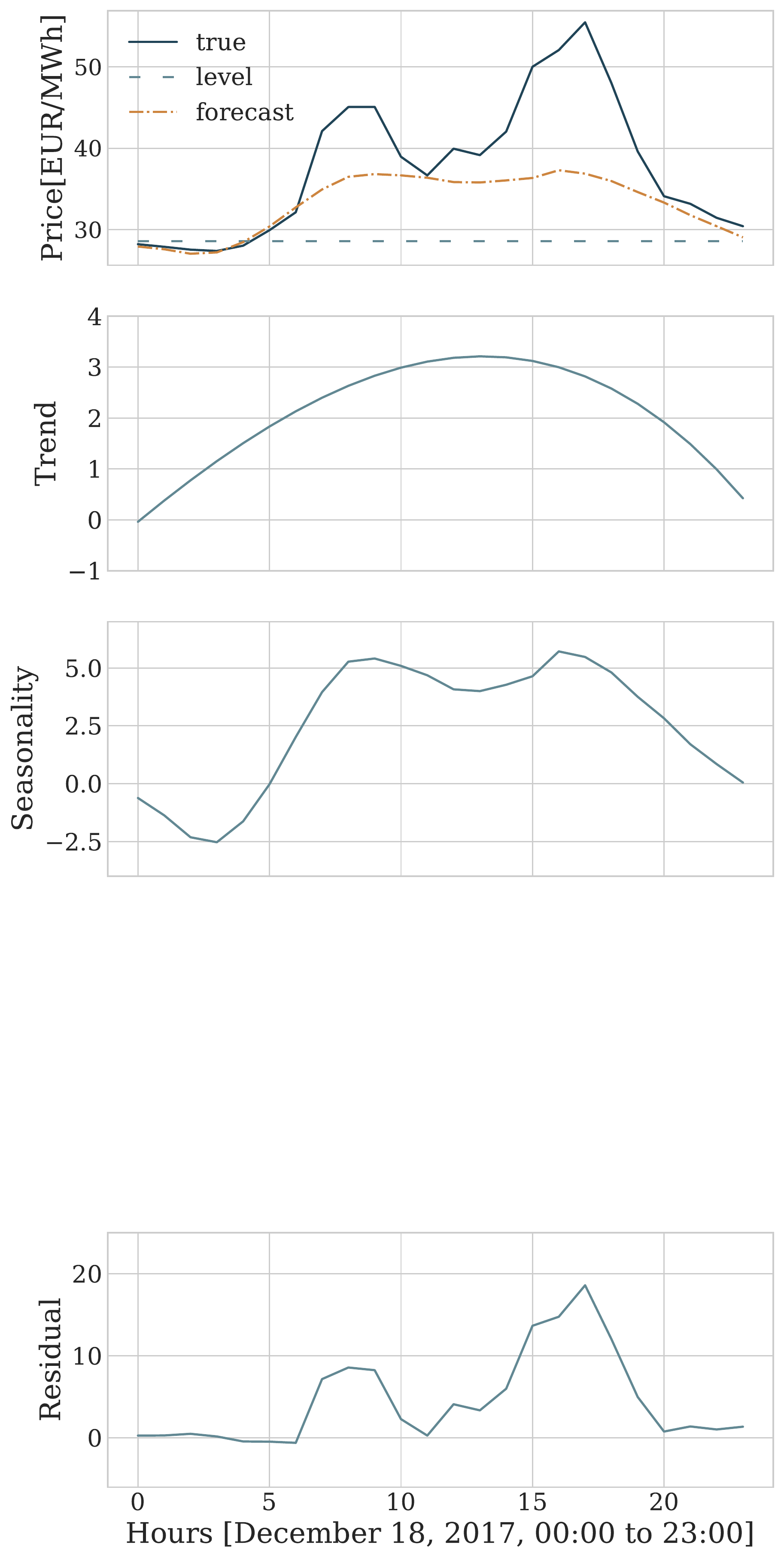}}
\subfigure[NBEATSx]{\label{fig:b}
\includegraphics[width=72mm]{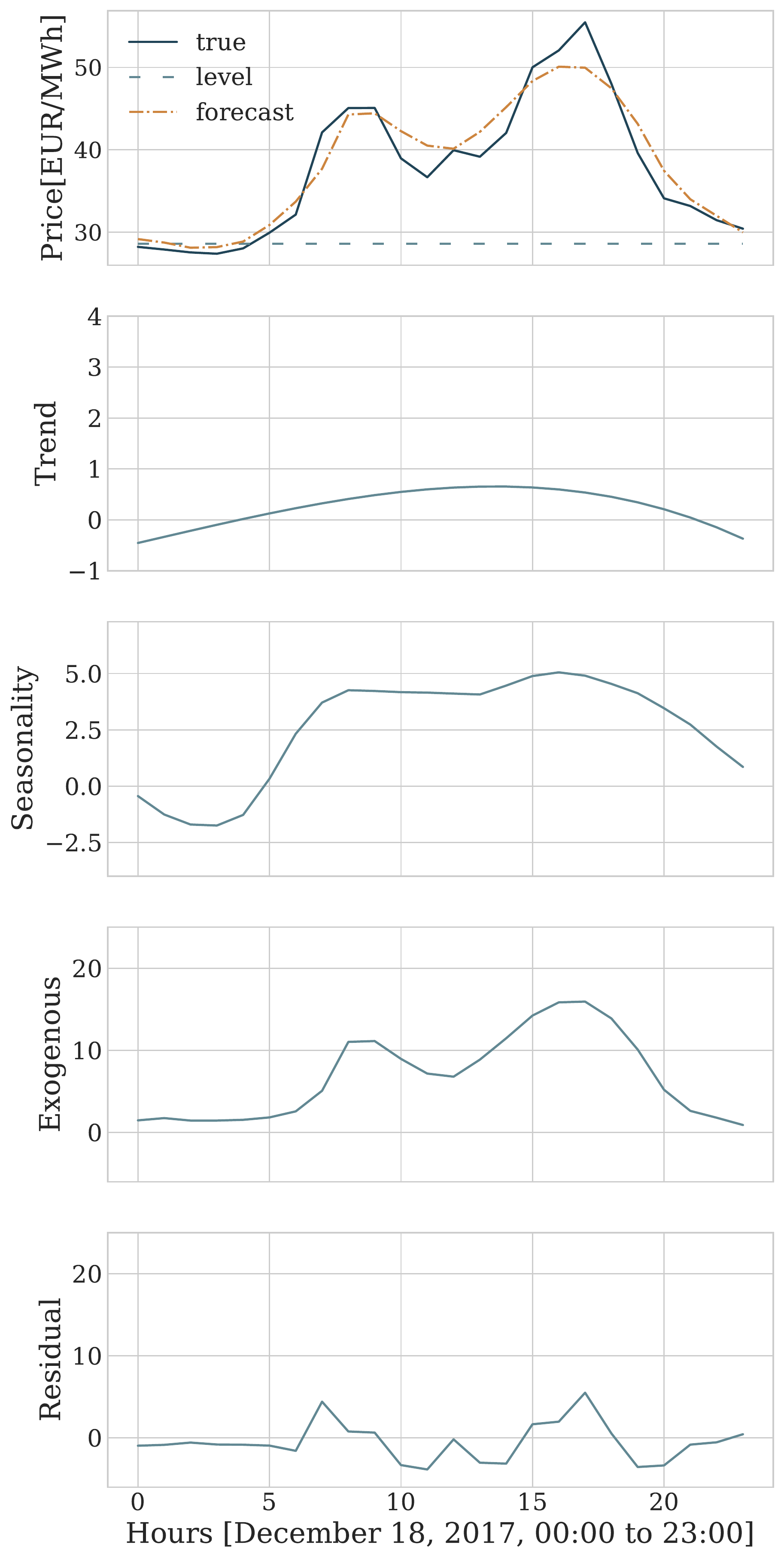}}
\caption{Time series signal decomposition for \NP\ electricity price day-ahead forecasts using interpretable variants of \NBEATS\ and \NBEATSx. 
The top row of graphs shows the original signal and the level, the latter is defined as the last available observation before the forecast. The second row shows the polynomial trend components, the third and fourth rows display the complex seasonality modeled by nonlinear Fourier projections and the exogenous effects of the electricity load on the price, respectively. The bottom row graphs show the unexplained variation of the signal. The use of electricity load and production forecasts turns out to be fundamental for accurate price forecasting.
}
\label{fig:forecast_decomposition}
\end{figure}

In this subsection, we demonstrate the versatility of the proposed method and show how a careful selection of the inductive bias, constituted by the assumptions used to learn the modeled signal, endows \NBEATSx\ with an outstanding ability to model complex dynamics while enabling human understanding of its outputs, turning it into a unique and exciting tool for time series analysis.
Our method combines the power of non-linear transformations provided by neural networks with the flexibility to model multiple seasons that can be fractional, and simultaneously account for interaction events such as holidays and other covariates. As described earlier, the interpretable configuration of the \NBEATSx\ architecture computes time-varying coefficients for slowly changing polynomial functions to model the trend, harmonic functions to model the cyclical behavior of the signal, and exogenous covariates. Here, we show how this configuration can decompose a time series into the classic set of level, trend, and seasonality components, while identifying the covariate effects. 
In this time series signal decomposition example, we show how the \NBEATSxi\ model benefits over \NBEATSi\ from explicitly accounting for information carried by exogenous covariates. Figure \ref{fig:forecast_decomposition} shows the \NP\ electricity market's hourly price (EUR/MWh), for December 18, 2017 which is a day with high prices due to high load. Other days have a less pronounced difference between the results obtained with the original \NBEATSi\ and the \NBEATSxi. We selected a day with a higher than normal load for exposition purposes, to demonstrate qualitative differences in the forecasts. We can see a substantial difference in the forecast residual magnitudes in the bottom row of Figure~\ref{fig:forecast_decomposition}.
\NBEATS\ shows a strong negative bias. 
On the other hand, \NBEATSxi\ is able to capture the evidently substantial explanatory value of the exogenous features, resulting in a much more accurate forecast.




\clearpage

\subsection{Comparative Analysis}
\subsubsection{Evaluation Metrics}

To ensure the comparability of our results with the existing literature, we opted to follow the widely accepted practice of evaluating the accuracy of point forecasts with the following metrics: 
\emph{mean absolute error} (MAE), \emph{relative mean absolute error} (rMAE)~\footnote{The naïve forecast method in EPF corresponds to a similar day rule, where the forecast for a Monday, Saturday and Sunday equals the value of the series observed on the same weekday of the previous week, while the forecast for Tuesday, Wednesday, Thursday, and Friday is the value observed  on the previous day.}, \emph{symmetric mean absolute percentage error} (sMAPE) and \emph{root mean squared error} (RMSE), defined as:
\begin{gather*}
\label{evaluation_metrics}
\begin{alignedat}{2}
MAE &= \frac{1}{24 N_{d}} \sum^{N_{d}}_{d=1}\sum^{24}_{h=1}|y_{d,h}-\hat{y}_{d,h}| 
&\quad\quad rMAE&=\frac{\sum^{N_{d}}_{d=1}\sum^{24}_{h=1}|y_{d,h}-\hat{y}_{d,h}|}{\sum^{N_{d}}_{d=1}\sum^{24}_{h=1}|y_{d,h}-\hat{y}^{naive}_{d,h}|} \\
&&\\
sMAPE &= \frac{200}{24 N_{d}} \sum^{N_{d}}_{d=1}\sum^{24}_{h=1} \frac{|y_{d,h}-\hat{y}_{d,h}|}{|y_{d,h}|+|\hat{y}_{d,h}|} 
&\quad\quad  RMSE &= \sqrt{\frac{1}{24 N_{d}} \sum^{N_{d}}_{d=1}\sum^{24}_{h=1} \left(y_{d,h}-\hat{y}_{d,h}\right)^{2}} \\
\end{alignedat}\\
\end{gather*}
where $y_{d,h}$ and $\hat{y}_{d,h}$ are the actual value and the forecast of the time series at day $d$ and hour $h$, \EDIT{for our experiments given the two years of each test set $N_{d}=728$}.

\EDITtwo{While regression-based models are estimated by minimizing squared errors, to train neural networks we minimize absolute errors (see Section \ref{section:training_methodology} below). Hence, both the MAE and RMSE are highly relevant in our context. Since they are not easily comparable across datasets -- and given the popularity of such errors in forecasting practice  \citep{makridakis2020m4_competition} -- we have additionally computed a percentage and a relative measure.} The sMAPE is used as an alternative to MAPE, which in the presence of values close to zero may degenerate \citep{hyndman2006another_look_measures}. 
\EDITtwo{The rMAE is calculated instead of a scaled measure used in the M4 competition for reasons explained in Sec.\ 5.4.2.\ of \cite{lago2021epftoolbox}.}

\subsubsection{Statistical Tests}

To assess which forecasting model provides better predictions, we rely on the \emph{Giacomini-White test} (GW; \citealt{giacomini_white2006test}) of the multi-step conditional predictive ability, which can be interpreted as a generalization of the \emph{Diebold-Mariano test} (DM; \citealt{diebold_mariano2002test}), widely used in the forecasting literature. Compared with the DM or other unconditional tests, the GW test is valid under general assumptions such as heterogeneity rather than stationarity of data. The GW test examines the null hypothesis of equal accuracy specified in Equation (\ref{equation:giacomini-white}), measured by the $L1$ norm of the daily errors of a pair of models $A$ and $B$, conditioned on the available information to that moment\footnote{In practice, the available information $\mathcal{F}_{d-1}$ is replaced with a constant and lags of the error difference $\Delta^{A,B}_{d}$ and the test is performed using a linear regression with a Wald-like test. When the conditional information considered is only the constant variable, one recovers the original DB test.} in time $\mathcal{F}_{d-1}$.

\begin{equation}
    H_{0}: \mathbb{E}\left[||\mathbf{y}_{d} - \hat{\mathbf{y}}^{A}_{d}||_{1} 
                         - ||\mathbf{y}_{d} - \hat{\mathbf{y}}^{B}_{d}||_{1} \;|\; \mathcal{F}_{d-1}\right]
           \equiv \mathbb{E}\left[ \Delta^{A,B}_{d} \;|\; \mathcal{F}_{d-1}\right] = 0
    \label{equation:giacomini-white}
\end{equation}

\subsubsection{Training Methodology}
\label{section:training_methodology}

The cornerstone of the training methodology for \NBEATSx\ and the benchmark models included in this work is the definition and use of the training, validation, early stopping, and test datasets depicted in Figure~\ref{fig:NP_electricity_datasets}.
The training set for each of the five markets comprises the first three years of data, the test set includes the last two years of data. The validation set is defined as the year between the train and test set coverages. The early stopping set, used for regularization, is either randomly sampled or corresponds to 42 weeks following the time span of the training set. These sets are used in the \emph{hyperparameter optimization phase} and \emph{recalibration phase} that we describe below. 

During the \emph{hyperparameter optimization phase}, model performance measured on the validation set is used to guide the exploration of the hyperparameter space defined in Table~\ref{table:hyperparameters}. 
During the \emph{recalibration phase}, the optimally selected model, as defined by its hyperparameters, is re-trained for each day to include newly available information before the test inference. In this phase, an early stopping set provides a regularization signal for the retraining optimization. 

To train the neural network, \EDITtwo{and as is common in the literature~\citep{smyl2020esrnn, oreshkin2020nbeats}}, we minimize the \emph{mean absolute error} (MAE) using stochastic gradient descent with \emph{Adaptive Moments} (\ADAM; \citealt{kingma2014method}). Figure~\ref{fig:training_curves} in the Appendix compares the training and validation trajectories for \NBEATS\ and \NBEATSx, as diagnostics to assess the differences of the methods. The early stopping strategy halts the training procedure if a specified number of consecutive iterations occur without improvements of the loss measured on the early stopping set~\citep{yuan2007early_stopping}. 

The \NBEATSx\ model is implemented and trained in \PyTorch~\citep{pytorch2019library} \  
and can be run with both CPU and GPU resources. The code is available publicly in a dedicated repository 
to promote reproducibility  of the presented results and to support related research.


\subsubsection{Hyperparameter Optimization}
\label{section:hyperparameter_optimization}

We follow the practice of \cite{lago2018DNN} to select the hyperparameters that define the model, input features, and optimization settings. During this phase, the validation dataset is used to guide the search for well performing configurations. To compare the benchmarks and the \NBEATSx, we rely on the same automated selection process: a Bayesian optimization technique that efficiently explores the hyperparameter space using tree-structured Parzen estimators~(\HYPEROPT; \citealt{bergstra2011hyperopt}). The architecture, optimization, and regularization hyperparameters are summarized in Table~\ref{table:hyperparameters}. To have comparable results, during the hyperparameter optimization stage we used the same number of configurations as in \cite{lago2018DNN}. Note, that some of the methods do not require any hyperparameter optimization -- e.g., the \ARone\ benchmark -- and some might only have one hyper-parameter to be determined, such as the regularization parameter in the \LEARx\ method, which is typically computed using the information criteria or cross-validation. 

\begin{table}[tbp]
\caption{Hyperparameters of \NBEATSx\ networks. They are common to all presented datasets. We list the typical values we considered in our experiments. The configuration that performed best on the validation set was selected automatically. 
} 
\label{table:hyperparameters}
\scriptsize
\centering
	\begin{tabular}{ll}
	\hline 
	\textsc{Hyperparameter} & \textsc{Considered Values}                                                                                                                \\ \hline
	\multicolumn{2}{c}{\textcolor{gray}{\textbf{{Architecture Parameters}}}}                                                                                            \\ \hline
	Input size, size of autorregresive feature window.                       & $L \in \{168\}$                                                                          \\
	Output size is the forecast horizon for day ahead forecasting.           & $H \in \{24\}$                                                                           \\
	List for architecture's type/number of stacks.                           & \{[\Identity, \TCN], [\TCN, \Identity]                                                   \\
	                                                                         & \qquad  [\Identity, \WaveNet], [\WaveNet, \Identity]\}                                   \\
	Type of activations used accross the network.                            & \{\SoftPlus,\SeLU,\PreLU,\Sigmoid,\ReLU, \TanH, \LReLU\}                                 \\
	Blocks separated by residual links per stack (shared across stacks).     &\{[1,1,1], [1, 1]\}.                                                                      \\
	\FCNN\ layers within each block.                                         &                          $\{2\}$    \\ 
	\FCNN\ hidden neurons on each layer of a block.                          &                          $\color{ForestGreen} N_{h}\in\{50,\dots,500\}$                  \\
	Exogenous Temp. convolution filter size (Equation~\ref{equation:exogenous_encoder})&                          $\{2,\dots,10\}$                                      \\
	Only interpretable, degree of trend polynomials.                         &                          $\color{ForestGreen} N_{pol}\in\{2,3,4\}$                       \\
	Only interpretable, number of Fourier basis (seasonality smoothness).    &                          $\color{ForestGreen} N_{hr}\in\{1,2\}$                          \\ 
	Whether \NBEATSx\ coefficients take input $\mathbf{X}$ (Equation~(\ref{equation:block_transformations})).
	                                                                         &                          $\{\text{True}, \text{False}\}$                                 \\ \hline
	\multicolumn{2}{c}{\textcolor{gray}{\textbf{{Optimization and Regularization parameters}}}}                                                                         \\ \hline
	Initialization strategy for network weights.                             & \{orthogonal, he\_norm, glorot\_norm\}                                                   \\
	Initial learning rate for regression problem.                            & Range(5e-4,1e-2)                                                                         \\
	The number of samples for each gradient step.                            & \{256, 512\}                                                                             \\
	The decay constant allows large initial lr to escape local minima.       & \{0.5\}                                                                                  \\
	Number of times the learning rate is halved during train.                & \{3\}                                                                                    \\
	Maximum number of gradient descent iterations.                           & \{30000\}                                                                                \\
	Iterations without validation loss improvement before stop.              & \{10\}                                                                                   \\
	Frequency of validation loss measurements.                               & \{100\}                                                                                  \\
	Whether batch normalization is applied after each activation.            & \{True, False\}                                                                          \\
	The probability for dropout of neurons in the projection layers.         & Range(0,1)                                                                               \\
	The probability for dropout of neurons for the exogenous encoder.        & Range(0,1)                                                                               \\
	Constant to control the Lasso penalty used on the coefficients.          & Range(0, 0.1)                                                                            \\
	Constant that controls the influence of L2 regularization of weights.    & Range(1e-5,1e-0)                                                                         \\
	The objective loss function with which \NBEATSx\ is trained.             & \{MAE\}                                                                                  \\
	Random weeks from full dataset used to validate.                         & \{42\}                                                                                   \\
	Number of iterations of hyperparameter search.                           & \{1500\}                                                                                 \\
	Random seed that controls initialization of weights.                     & DiscreteRange(1,1000)                                                                    \\ \hline
	\multicolumn{2}{c}{\textcolor{gray}{\textbf{{Data Parameters}}}}                                                                                                    \\ \hline
	Rolling window sample frequency, for data augmentation.                  & \{1, 24\}                                                                                \\
	Number of time windows included in the full dataset.                     & 4 years                                                                                  \\
	Number of validation weeks used for early stopping strategy.             & \{40, 52\}                                                                               \\
	Normalization strategy of model inputs.                                  & \{none, median, invariant, std \}                                                        \\ \hline
	\end{tabular}
\end{table}


\subsubsection{Ensembling}
\label{section:ensembling}

In many recent forecasting competitions, and particularly in the M4  competition, most of the top-performing models were ensembles~\citep{atiya2019combinations}. It has been shown that in practice, combining a diverse group of models can be a powerful form of regularization to reduce the variance of predictions~\citep{breiman1996bagging,nowotarski2014ensembles, hubicka2019ensembles}.

The techniques used by the forecasting community to induce diversity in the models are plentiful. The original \NBEATS\ model obtained its diversity from three sources, training with different loss functions, varying the size of the input windows, and bagging models with different random initializations \citep{oreshkin2020nbeats}. They used the median as the aggregation function for 180 different models. Interestingly, the original model did not rely on regularization, such as L2 or dropout, as \cite{oreshkin2020nbeats} found it to be good for the individual models but detrimental to the ensemble. 

In our case, we ensemble the \NBEATSx\ model using two sources of diversity. The first being a data augmentation technique controlled by the sampling frequency of the windows used during training, as defined in the data parameters from Table~\ref{table:hyperparameters}. The second source of diversity being whether we randomly select the early stopping set or instead use the last 42 weeks preceding the test set. Combining the data augmentation and early stopping options, we obtain four models that we ensemble using arithmetic mean as the aggregation function. This technique is also used by the \DNN\ benchmark \citep{lago2018DNN, lago2021epftoolbox}.
\subsubsection{Forecasting Results}
\label{subsection:forecasting_results}

We conducted an empirical study involving two types of \emph{Autoregressive Models} (\ARone \ and \ARxone; \citealt{weron2014epf_survey1}), the \emph{Lasso Estimated Auto-Regressive} (\LEARx; \citealt{weron2016LEAR}), a parsimonious \emph{Deep Neural Network} (\DNN; \citealt{lago2018DNN, lago2021epftoolbox}), the original \emph{Neural Basis Expansion Analysis} without exogenous covariates (\NBEATS; \citealt{oreshkin2020nbeats}), and the \emph{Exponential Smoothing Recurrent Neural Network} (\ESRNN; \citealt{smyl2020esrnn}).
This experiment examines the effects of including the covariate inputs and comparing \NBEATSx\ with state-of-the-art methods for the electricity price day-ahead forecasting task.

Table~\ref{table:main_results_ensemble} summarizes the performance of the ensembled models where \NBEATSx\ ensemble shows prevailing performance. It improves 18.77\% on average for all metrics and markets when compared with the original \NBEATS\ and 20.6\% when compared to \ESRNN\ without time-dependent covariates. For the ensembled models, \NBEATSx\ RMSE improved on average 4.68\%, MAE improved 2.53\%, rMAE improved 1.97\%,and sMAPE improved 1.25\%. When comparing \NBEATSx\ ensemble against \DNN\ ensemble on individual markets, \NBEATSx\ improved by 5.38\% on the NordPool market, by 2.48\% on French market and 2.81\% on German market. There was a non-significant difference of \NBEATSx\ performance on \PJM\ and \BE\ markets of 0.24\% and 1.1\%, respectively.

\begin{figure}[tbp]
\centering
\includegraphics[width=.3\textwidth]{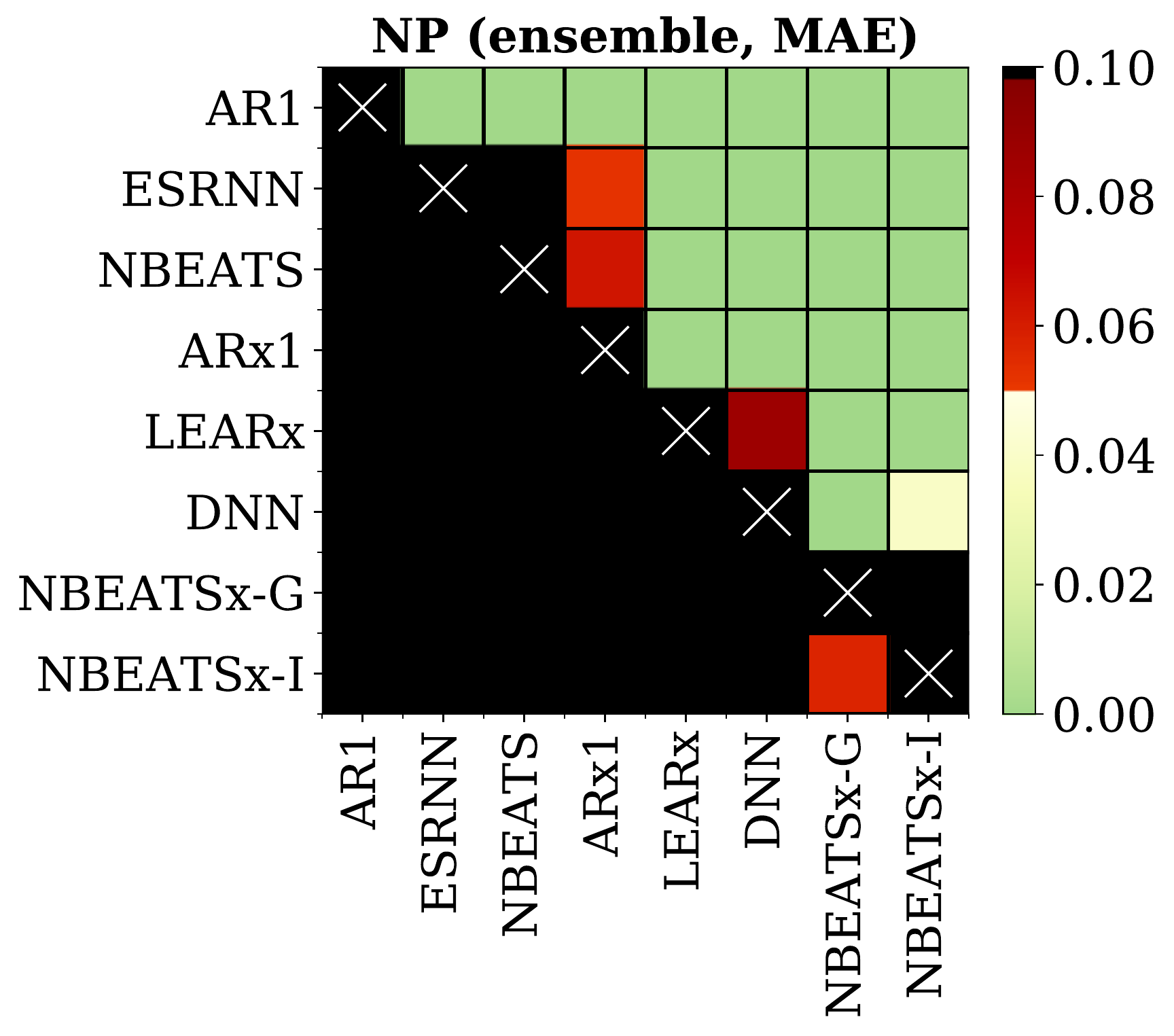}\quad
\includegraphics[width=.3\textwidth]{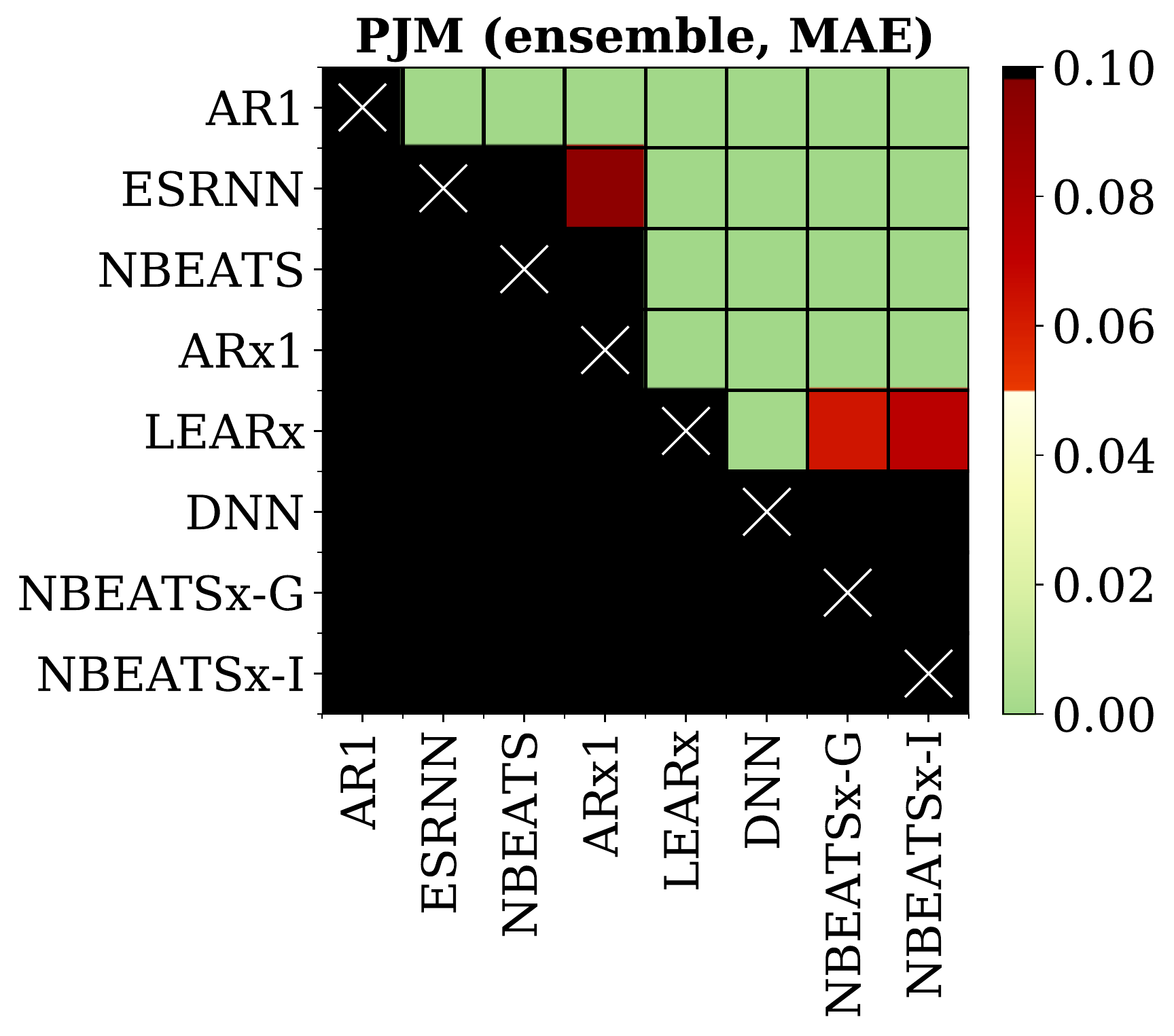}\quad
\includegraphics[width=.3\textwidth]{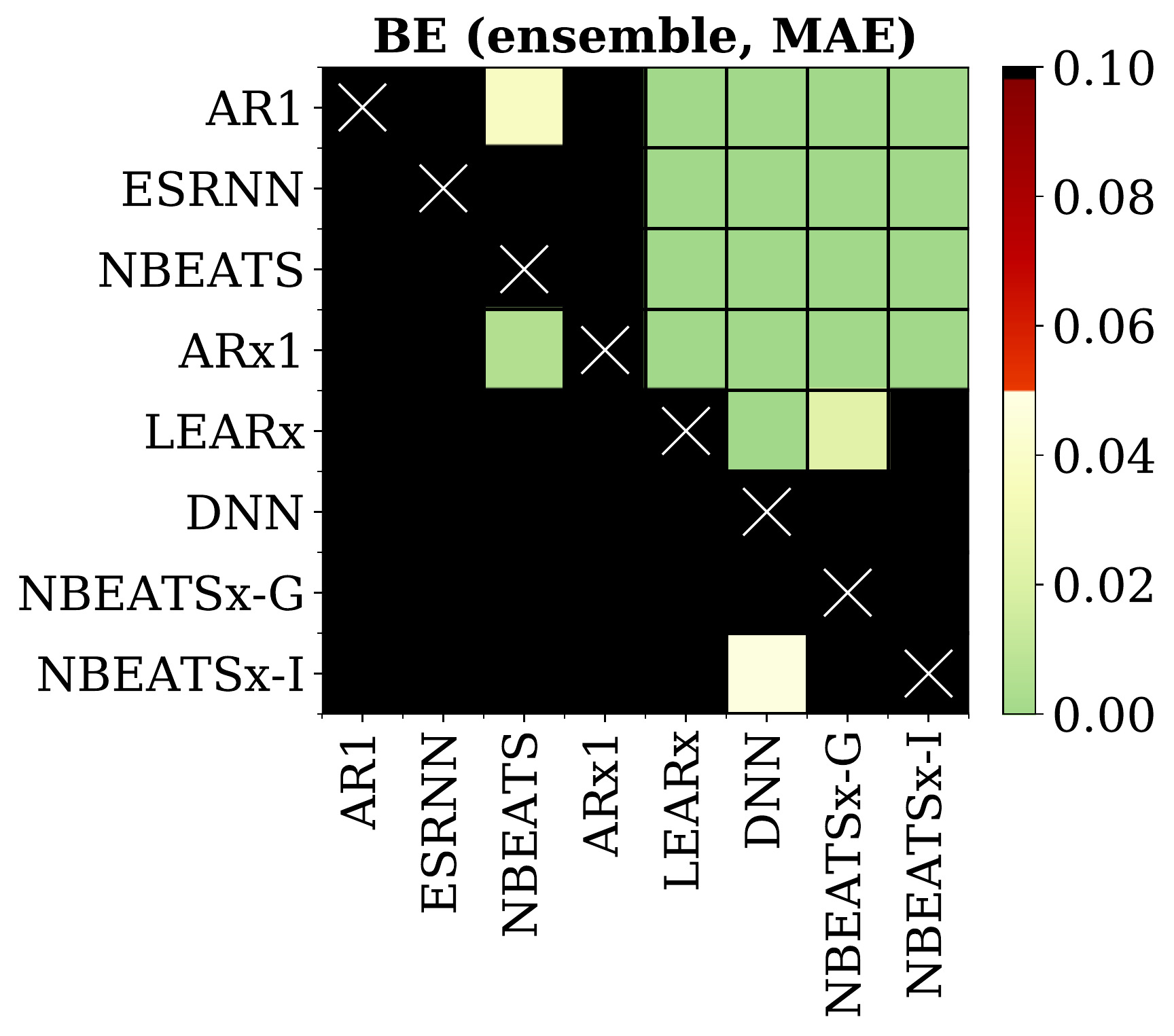}
\medskip
\includegraphics[width=.3\textwidth]{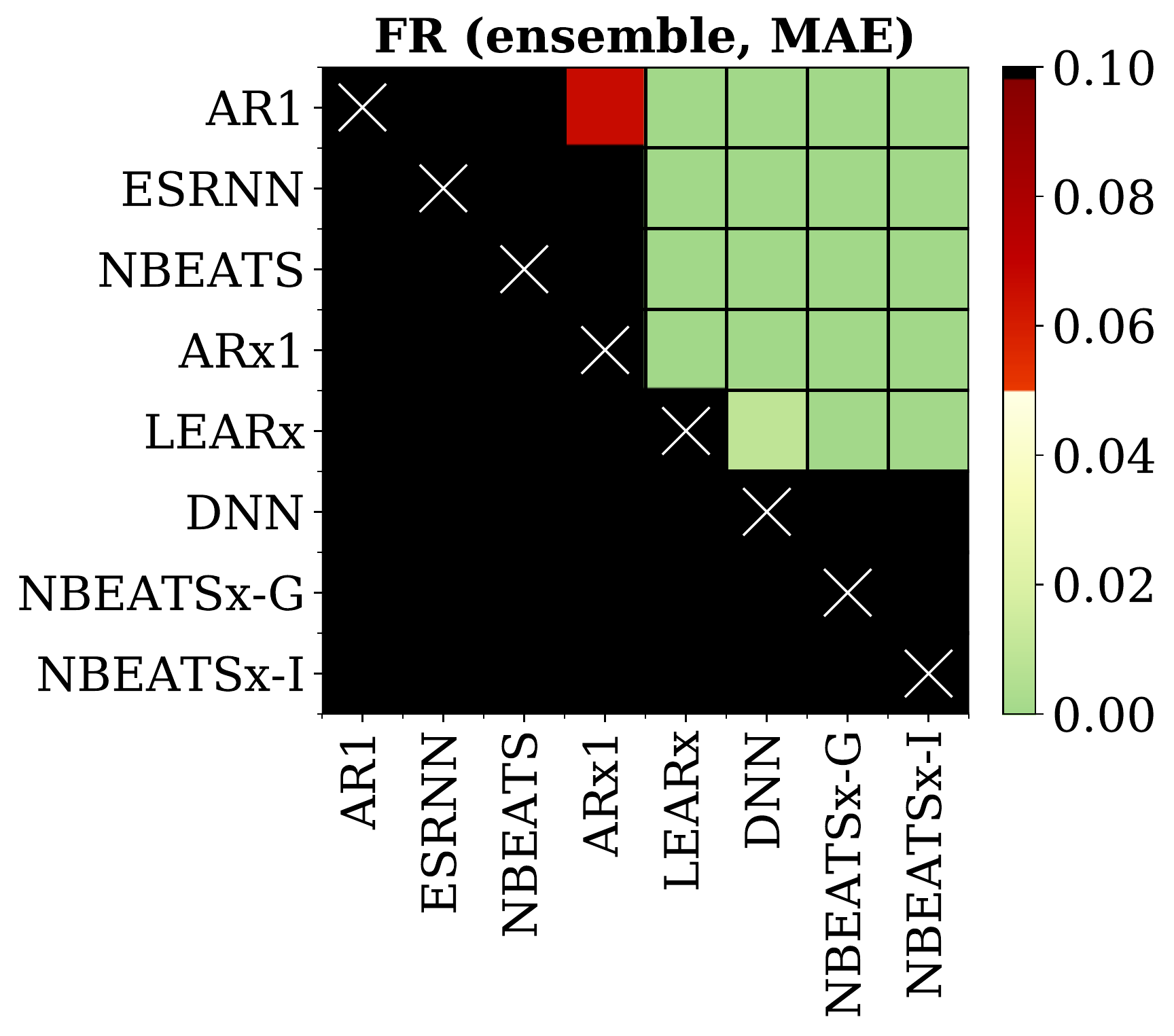}\quad
\includegraphics[width=.3\textwidth]{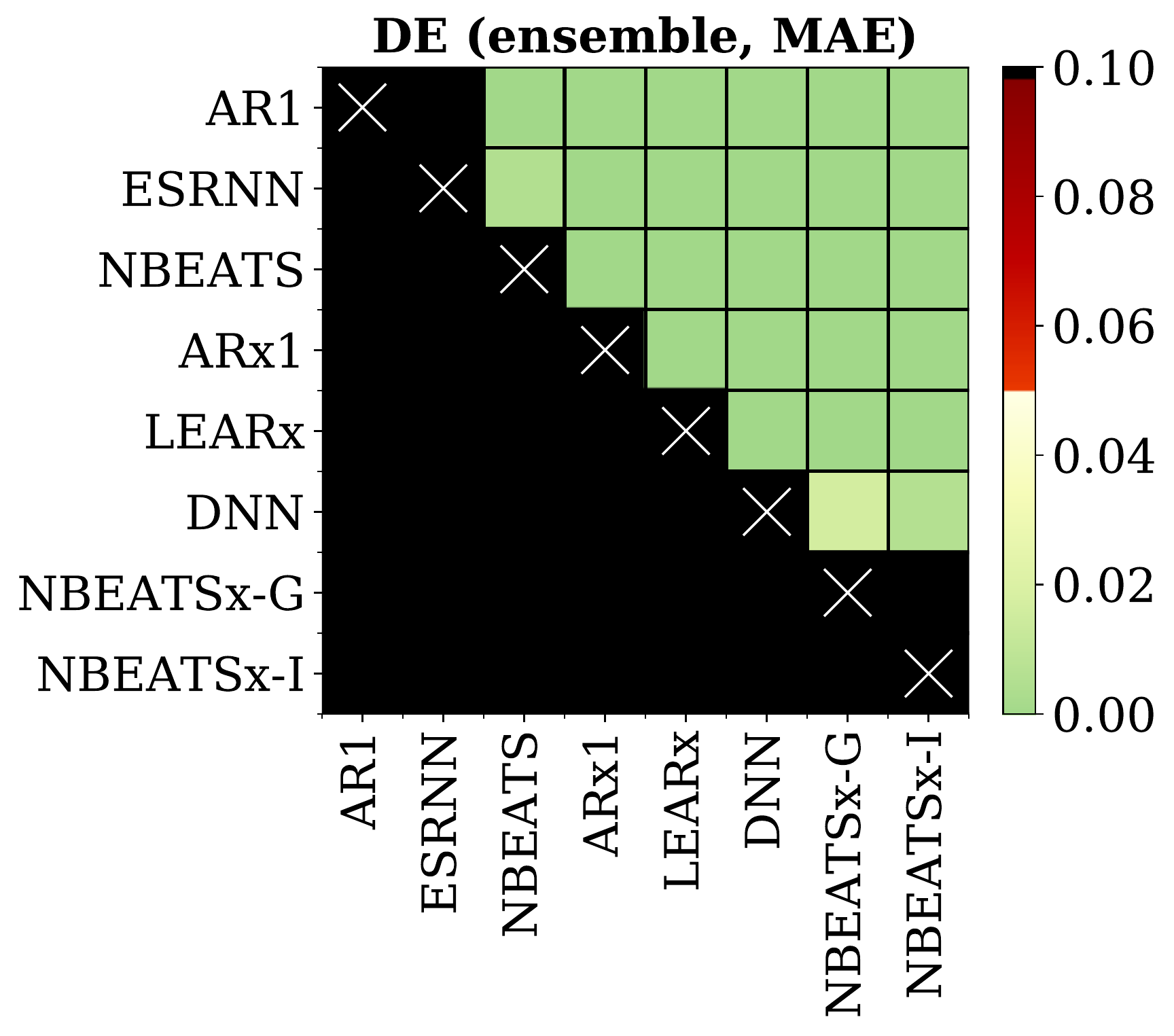}
\caption{Results of the Giacomini-White test for the day-ahead predictions with \emph{mean absolute error} (MAE) applied to pairs of the ensembled models on the five electricity markets datasets. Each grid represents one market. Each colored cell in a grid is plotted black, unless the predictions of the model corresponding to its column of the grid outperforms the predictions of the model corresponding to its row of the grid.
The color scale reflects significance of the difference in MAE, with solid green representing the lowest $p$-values.
} \label{fig:pvals_ensembleMAE}
\end{figure}

Figure~\ref{fig:pvals_ensembleMAE} provides a graphical representation of the statistical significance from the \emph{Giacomini-White} test (GW) for the six ensembled models, across the five markets for the MAE evaluation metric. A similar significance analysis is conducted for the single models. The models included in the significance tests are the same as in Table~\ref{table:main_results_ensemble}: \LEAR, \DNN, \ESRNN, \NBEATS, and our proposed methods, \NBEATSxg\ and \NBEATSxi. The $p$-value of each comparison shows if the performance improvement of the model's predictions corresponding to the column index of a cell in the grids shown in Figure~\ref{fig:pvals_ensembleMAE} over the model's predictions corresponding to the row of this cell of the grid is statistically significant.
\NBEATSxg\ model outperformed \DNN\ model in \NP\ and \DE, while \NBEATSxi\ outperformed it in \NP, \FR, and \DE. Moreover, no benchmark model significantly outperformed \NBEATSxi\ and \NBEATSxg\ in any market.

In the \ref{section:appendix} we observe similar results for the single best models chosen from the four possible configurations of the ensemble components described in Section \ref{section:ensembling}.
Table~\ref{table:main_results_single} summarizes the accuracy of the predictions measured with the MAE and Figure \ref{fig:pvals_singleMAE} displays the significance of the GW test. Ensembling improves the accuracy of \NBEATSx\ by 3\% on average acrosss all markets, when compared to the single best models.

Finally, regarding the computational time complexity \NBEATSx\ maintains good performance. As shown in Table~\ref{table:computational_time} in the Appendix, the time necessary to compute day-ahead predictions is in the order of miliseconds and comparable to that of the \LEAR\ and \DNN\ benchmarks. Additionally, the average time needed to perform a recalibration only takes circa 50 percent more than the relatively parsimonious \DNN. 


\begin{table}[tbp]
\caption{Forecast accuracy measures for day-ahead electricity price predictions of \emph{ensembled models}. The \ESRNN \ and \NBEATS \ do not include time dependent covariates. The reported metrics are \emph{mean absolute error} (MAE), \emph{relative mean absolute error} (rMAE), \emph{symmetric mean absolute percentage error} (sMAPE) and \emph{root mean squared error} (RMSE).  The smallest errors in each row are highlighted in bold.
\\ \textsuperscript{*} The \LEARx \ results for \DE \ differ from \cite{lago2021epftoolbox} -- the values presented there are revised \citep{lago2021erratum}}. 
\label{table:main_results_ensemble}
\centering
\scriptsize
\begin{tabular}{llcccccccc}
                         &       & \ARone     & \ESRNN  & \NBEATS & \ARxone & \LEARx\textsuperscript{*} & \DNN            & \NBEATSxg           & \NBEATSxi     \\ \hline
\multirow{4}{*}{\NP}     & MAE   &   2.26   &  2.09   &  2.08   &   2.01    &   1.74                    & 1.68            & \textbf{1.58}       & 1.62          \\
                         & rMAE  &   0.71   &  0.66   &  0.66   &   0.63    &   0.55                    & 0.53            & \textbf{0.50}       & 0.51          \\
                         & sMAPE &   6.47   &  6.04   &  5.96   &   5.84    &   5.01                    & 4.88            & \textbf{4.63}       & 4.70          \\
                         & RMSE  &   4.08   &  3.89   &  3.94   &   3.71    &   3.36                    & 3.32            & \textbf{3.16}       & 3.27          \\ \hline
\multirow{4}{*}{\PJM}    & MAE   &   3.83   & 3.59    &  3.49   &   3.53    &   3.01                    & \textbf{2.86}   & 2.91                & 2.90          \\
                         & rMAE  &   0.79   &  0.74   &  0.72   &   0.73    &   0.62                    & \textbf{0.59}   & 0.60                & 0.60          \\
                         & sMAPE &   14.5   &  14.12  & 13.57   &   13.64   &  11.98                    & \textbf{11.33}  & 11.54               & 11.61         \\
                         & RMSE  &   6.24   &  5.83   &  5.64   &   5.74    &   5.13                    & 5.04            & 5.02                & \textbf{4.84} \\ \hline
\multirow{4}{*}{\BE}     & MAE   &   7.2    &  6.96   &  6.84   &   7.19    &   6.14                    & \textbf{5.87}   & 5.95                & 6.11           \\
                         & rMAE  &   0.88   &  0.85   &  0.83   &   0.88    &   0.75                    & \textbf{0.72}   & 0.73                & 0.75           \\
                         & sMAPE &   16.26  &  15.84  & 15.80   &   16.11   &  14.55                    & \textbf{13.45}  & 13.86               & 14.02          \\
                         & RMSE  &   18.62  &  16.84  & 17.13   &   18.07   &  15.97                    & 15.97           & \textbf{15.76}      & 15.80          \\ \hline
\multirow{4}{*}{\FR}     & MAE   &   4.65   &  4.65   &  4.74   &   4.56    &   3.98                    &  3.87           & 3.81                & \textbf{3.79}  \\
                         & rMAE  &   0.78   &  0.78   &  0.80   &   0.76    &   0.67                    & 0.65            & \textbf{0.64}       & \textbf{0.64}  \\
                         & sMAPE &   13.03  &  13.22  & 13.30   &   12.7    &  11.57                    & 10.81           & \textbf{10.59}      & 10.69          \\
                         & RMSE  &   13.89  & 11.83   & 12.01   &   12.94   &  \textbf{10.68}           & 11.87           & 11.50               & 11.25          \\ \hline
\multirow{4}{*}{\DE}     & MAE   &   5.74   & 5.60 & 5.31       &   4.36    &   3.61                    &  3.41           & 3.31                & \textbf{3.29}  \\
                         & rMAE  &   0.71   & 0.70 & 0.66       &   0.54    &   0.45                    & 0.42            & \textbf{0.41}       & \textbf{0.41}  \\
                         & sMAPE &   21.37  & 20.97 & 19.61     &   17.73   &  14.74                    & 14.08           & \textbf{13.99}      & \textbf{13.99} \\
                         & RMSE  &   9.63   & 9.09 & 8.99       &   7.38    &   6.51                    &  5.93           & 5.72                & \textbf{5.65}  \\ \hline
\end{tabular}
\end{table}

\clearpage


\section{Conclusions} \label{section:conclusion}
We have presented \NBEATSx: the new method for univariate time series forecasting with exogenous variables.
It extends the well-performing neural basis expansion analysis. The resulting neural based method has several valuable properties that make it suitable for a wide range of forecasting tasks. 
The network is fast to optimize as it is mainly composed of fully-connected layers. It can produce interpretable  results, and achieves state-of-the-art performance on forecasting tasks where consideration of exogenous variables is fundamental.

We demonstrated the utility of the proposed method using a set of benchmark datasets from electricity price forecasting domain, but it can be straightforwardly applied to forecasting problems in other domains.
Qualitative evaluation shows that the interpretable configuration of \NBEATSx\ can provide valuable insights to the analyst, as it explains the variation of the time series by separating it into trend, seasonality, and exogenous components, in a fashion analogous to classic time series decomposition. Regarding the quantitative forecasting performance, we observed no significant differences between \ESRNN\ and \NBEATS\ without exogenous variables. At the same time, \NBEATSx\ improves over \NBEATS\ by nearly 20\%  and up to 5\% over \LEAR\ and \DNN\ models  specialized for the Electricity Price Forecasting tasks. 
Finally, we found no significant trade-offs between the accuracy and interpretability of \NBEATSxg\ and \NBEATSxi\ predictions.

The neural basis expansion analysis is a very 
flexible method capable of producing accurate
and interpretable forecasts, yet there is still room for improvement. For instance, augmentation of the harmonic functions towards wavelets or replacement of the convolutional encoder that would generate the covariate basis with smoothing alternatives such as splines. Additionally, one can extend the current non-interpretable method by regularizing its outputs with smoothness constraints.



\section*{Acknowledgements} \label{section:acknowledgements}
This work was partially supported by the Defense Advanced Research Projects Agency (award FA8750-17-2-0130), the National Science Foundation (grant 2038612), the Space Technology Research Institutes grant from NASA’s Space Technology Research Grants Program, 
the U.S.\ Department of Homeland Security (award 18DN-ARI-00031),
the Ministry of \EDIT{Education and Science (MEiN,} Poland; grant 0219/DIA/2019/48),
the National Science Center (NCN, Poland; grant 2018/30/A/HS4/00444), and Nixtla. Kin G.\ Olivares and Cristian Challu want to thank Stefania La Vattiata, Max Mergenthaler and Federico Garza for their support. 

\bibliography{citations.bib}
\bibliographystyle{model5-names}

\clearpage
\appendix
\section{Appendix} \label{section:appendix}
\setcounter{table}{0}
\setcounter{figure}{0}
\renewcommand{\thetable}{A\arabic{table}}

\subsection{Forecast and Backast Basis}
\label{appendix:basis}

\begin{figure}[h] 
\centering
\subfigure[Trend Basis]{\label{fig:trend_basis}
\includegraphics[width=160mm]{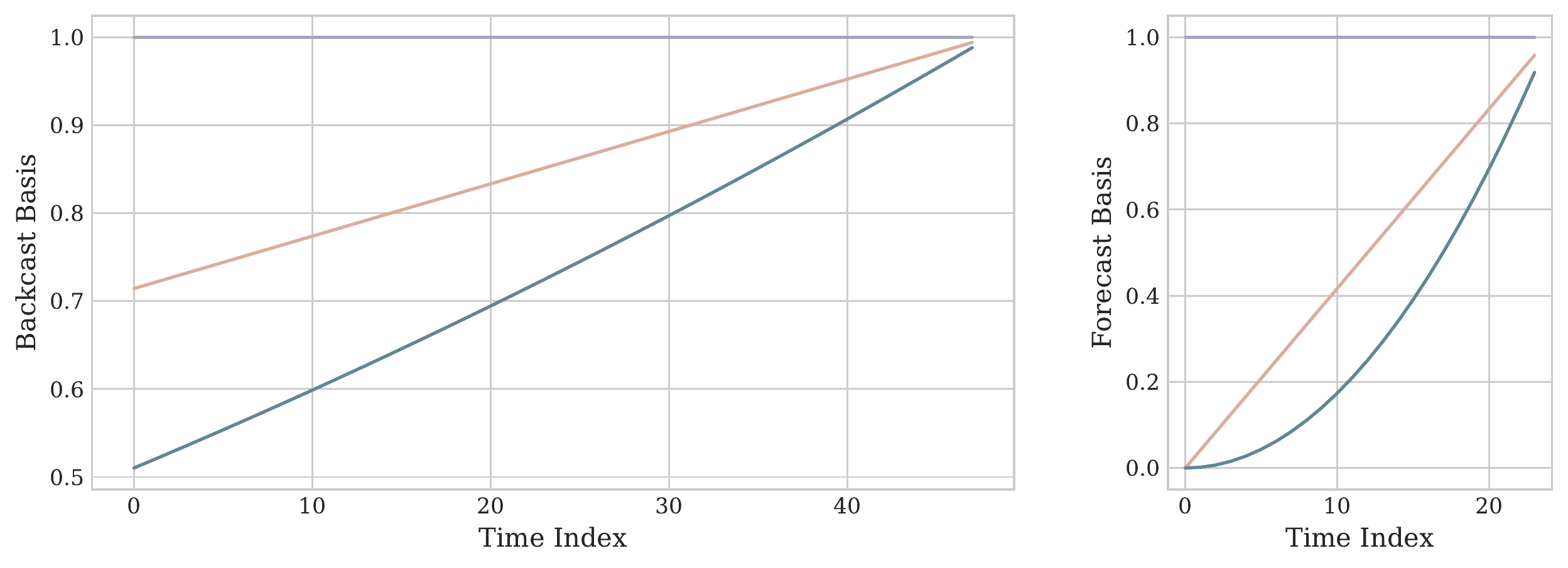}} 
\subfigure[Harmonic Basis]{\label{fig:harmonic_basis}
\includegraphics[width=160mm]{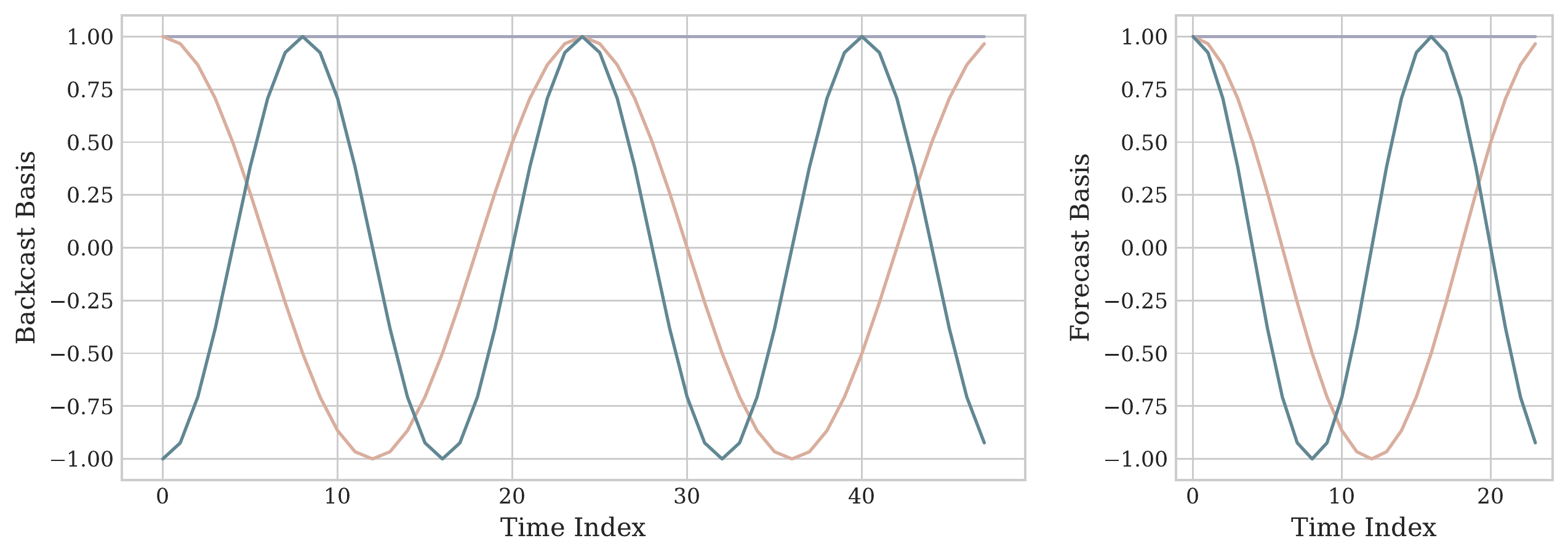}}
\caption{Examples of polynomial and harmonic basis included in the interpretable configuration of the neural basis expansion analysis. The slowly varying basis allow \NBEATS\ to model trends and seasonalities.}
\label{fig:forecast_backckast_basis}
\end{figure}

As discussed in Section~\ref{subsection:nbeatsx_configurations}, the interpretable configuration of the \NBEATSx\ method performs basis projections into polynomial functions for the trends, harmonic functions for the seasonalities and exogenous variables. As shown in Figure~\ref{fig:forecast_backckast_basis}, both the forecast and the backcast components of the model rely on similar basis functions, and the only difference depends upon the span of their time indexes. For this work in the EPF application of \NBEATS, the backcast horizon corresponds to 168 hours while the forecast horizon corresponds to 24.

\clearpage
\subsection{Training and validation curves}

\begin{figure}[h] 
\centering
\subfigure[Train set]{\label{fig:train_curve}
\includegraphics[width=120mm]{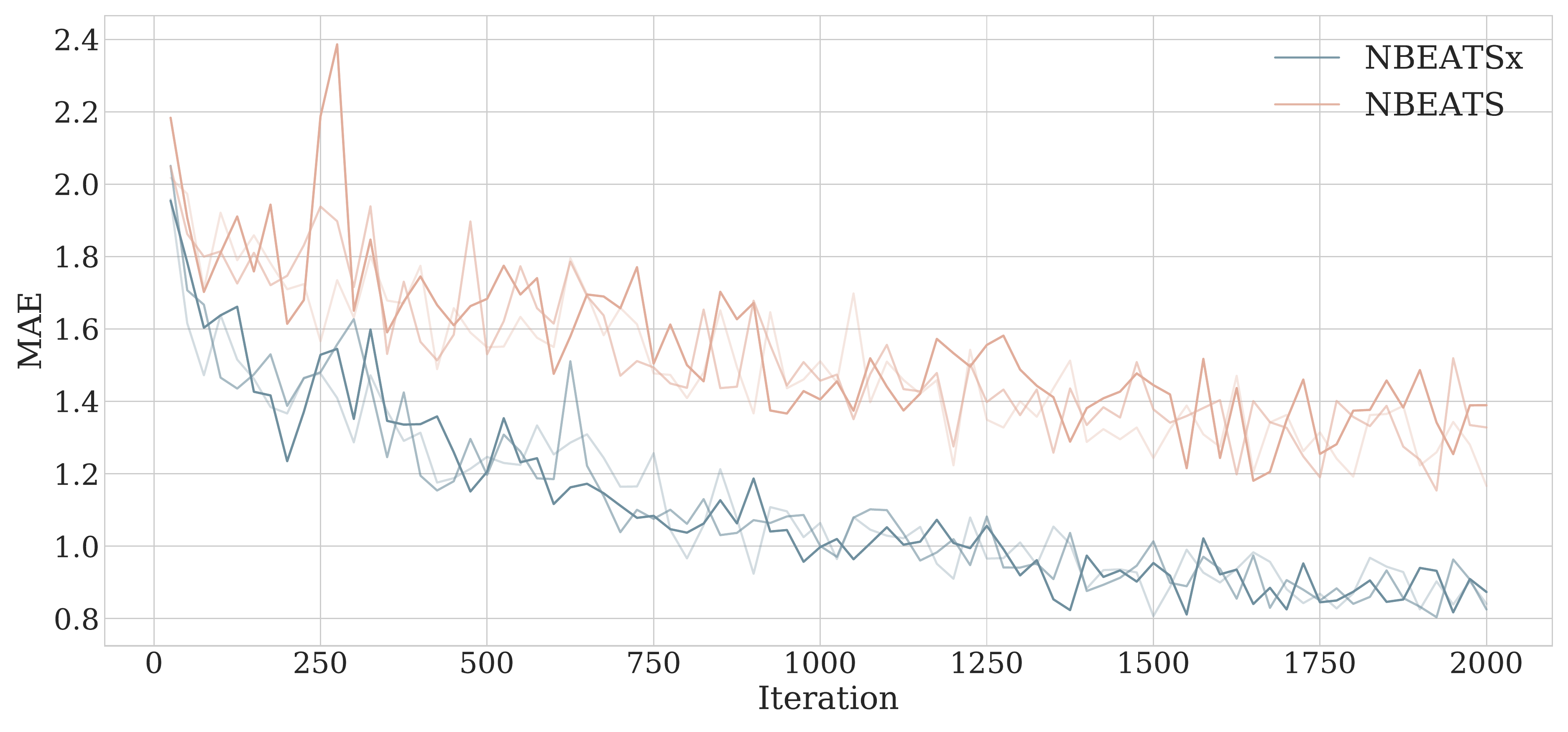}}
\subfigure[Validation set]{\label{fig:validation_curve}
\includegraphics[width=120mm]{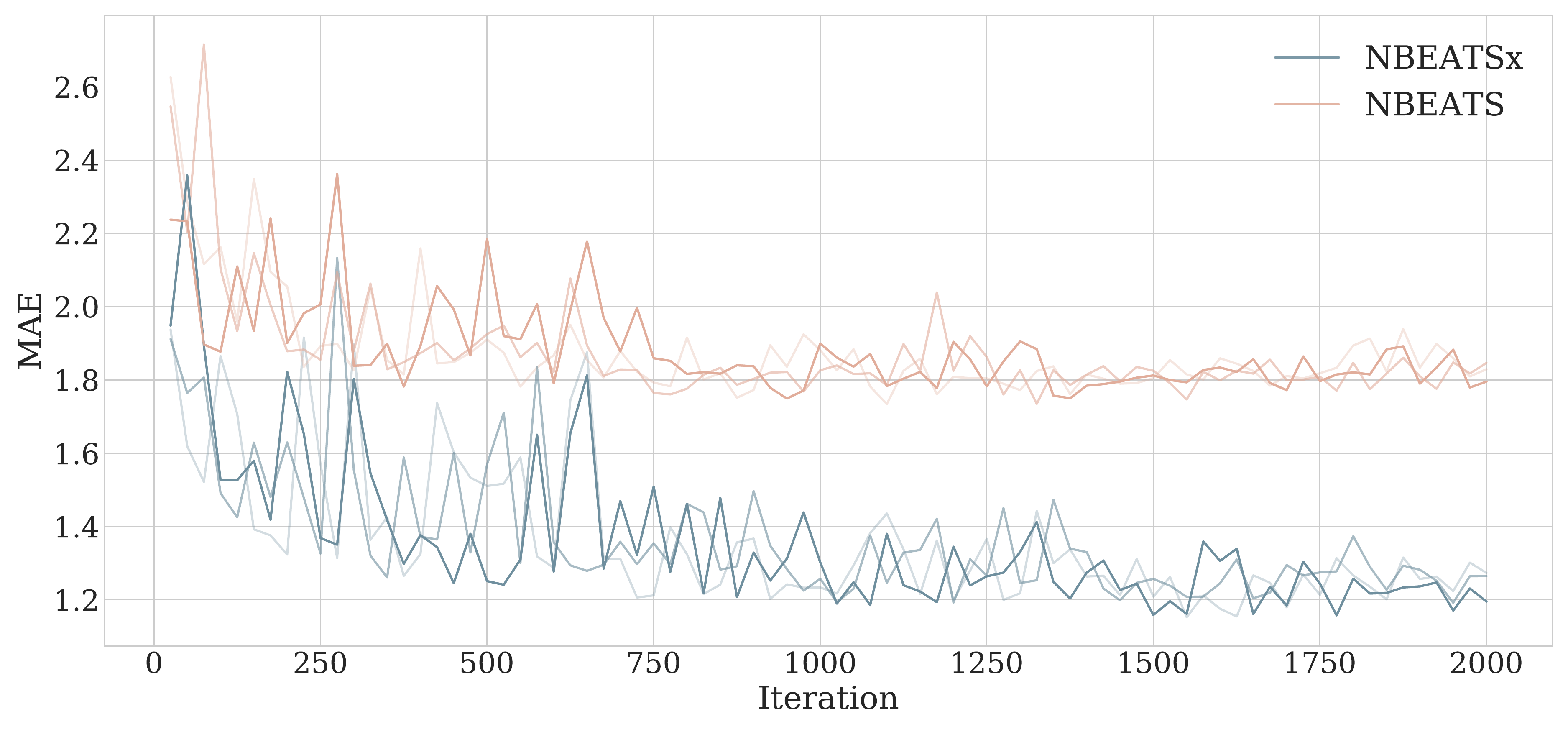}} 
\caption{Training and validation \emph{Mean Absolute Error} (MAE) curves on the \NP\ market. We show the curves for \NBEATSxg\ with exogenous variables and \NBEATS\ without exogenous variables as a function of the optimization iterations. We define the four curves by a different random seed used for initialization.
}
\label{fig:training_curves}
\end{figure}

To study the effects of exogenous variables on the \NBEATS\ model, we performed model training procedure diagnostics.
Figure~\ref{fig:training_curves} shows the train and validation \emph{mean absolute error} (MAE) for the \NBEATS\ and \NBEATSx\ models as training progresses. The curves correspond to the hyperparameter optimization phase described in Section \ref{section:hyperparameter_optimization}. The models trained with and without exogenous variables display a considerable difference in their train and validation errors as observed by the two separate clusters of trajectories. The exogenous variables, in this case, the electricity load and production forecasts, significantly improve the neural basis expansion analysis.

\clearpage

\subsection{Computational Time}
\label{appendix:computation}

\begin{table}[h]
\caption{Computational time performance in seconds for the top four most accurate models for the day-ahead electricity price forecasting task in the \NP\ market, averaged for the four elements of the ensembles (Time performance for the rest of the markets is almost identical).}
\label{table:computational_time}
\centering
\begin{tabular}{lcccc}
\toprule
              & \LEARx  & \DNN     & \NBEATSxg & \NBEATSxi \\
\midrule
Recalibration & 18.57   & 50.65    & 75.02     & 81.61    \\
Prediction    & 0.0032  & 0.0041   & 0.0048    & 0.0054   \\
\midrule
\end{tabular}
\end{table}

We measured the computational time of the top four best algorithms with two metrics: the recalibration of the ensemble models selected from the hyperparameter optimization, and the computation of the predictions. For these experiments, we used a GeForce RTX 2080 GPU for the neural network models and an Intel(R) Xeon(R) Silver 4210 CPU @ 2.20GHz for \LEAR. 

The training time of the \emph{recalibration phase} of \NBEATSx\ remains efficient, as it still trains in 75 and 81 seconds, increasing by 30 seconds on the relatively simple \DNN. The computational time of the prediction remains within miliseconds. Finally the \emph{hyperparameter optimization} scales linearly with respect to the time of the \emph{recalibration phase} and the evaluation steps of the optimization, in case of the \NBEATSxg\ the approximate time of a hyperparameter search of 1000 steps takes two days\footnote{For comparability we use 1000 steps \citep{lago2021epftoolbox}, restricting to 300 steps yields similar results.}.


\subsection{Best Single Models}
Table~\ref{table:main_results_single} shows that the best \NBEATSx\ models yield improvements of 14.8\% on average across all the evaluation metrics when compared to its \NBEATS\ counterpart without exogenous covariates, and improvements of 23.9\% when compared to \ESRNN\ without time-dependent covariates. A perhaps more remarkable result is the statistically significant improvement of forecast accuracy over \LEAR\ and \DNN\ benchmarks, ranging from 0.75\% to 7.2\% across all metrics and markets, with the exception of \BE.
Compared to \DNN, the RMSE improved on average 4.9\%, the MAE improved 3.2\%, the rMAE improved 3.0\%, and sMAPE improved 1.7\%. When comparing the best \NBEATSx\ models against the best \DNN\ on individual markets, \NBEATSx\ improved by 3.18\% on the Nord Pool market (\NP), 2.03\% 2.65\% on French (\FR) and  5.24\% on German (\DE) power markets. 
The positive difference in performance for Belgian (\BE) market of 0.53\% was not statistically significant. 

\begin{figure}[bp]
\centering
\includegraphics[width=.28\textwidth]{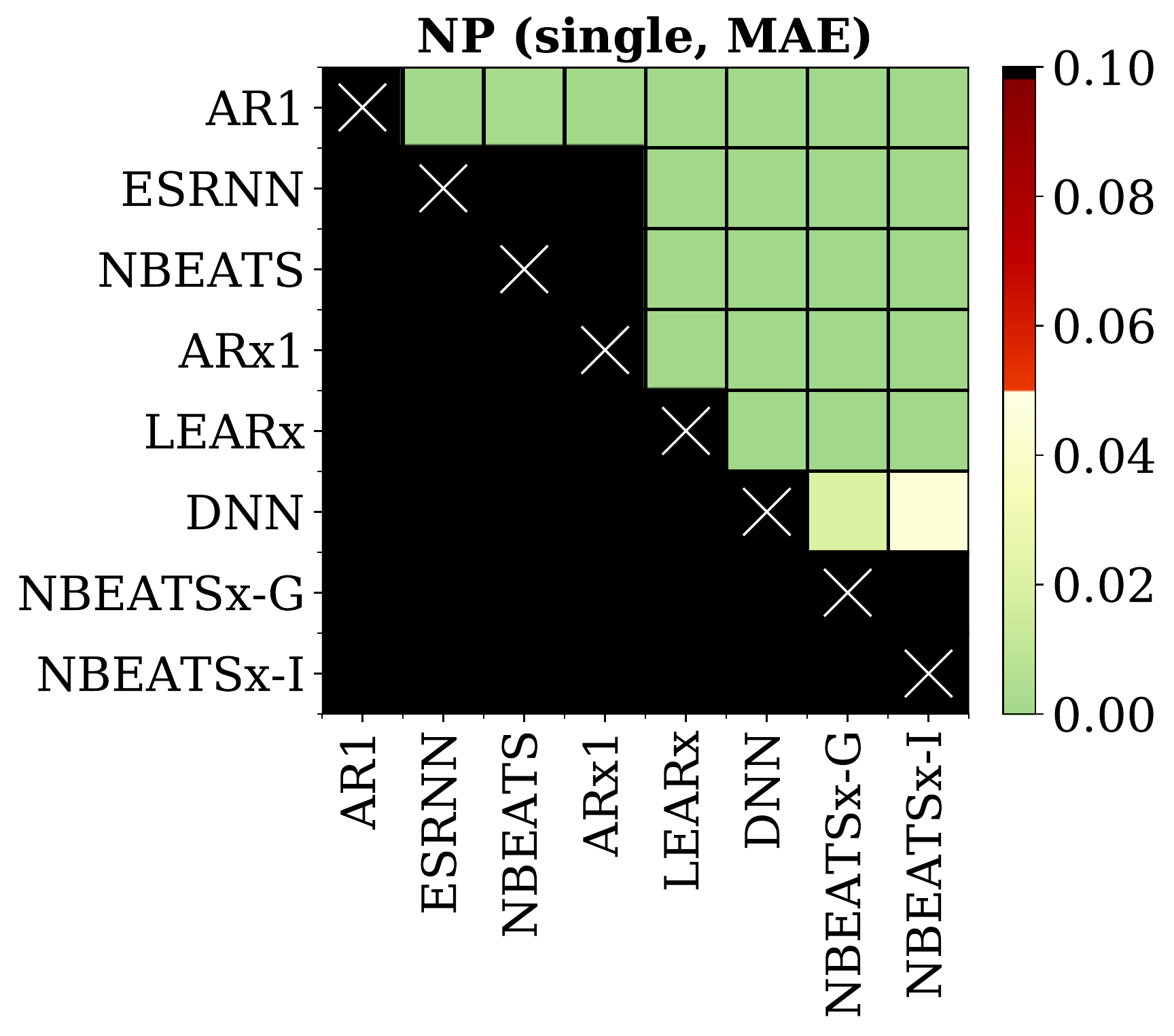}\quad
\includegraphics[width=.28\textwidth]{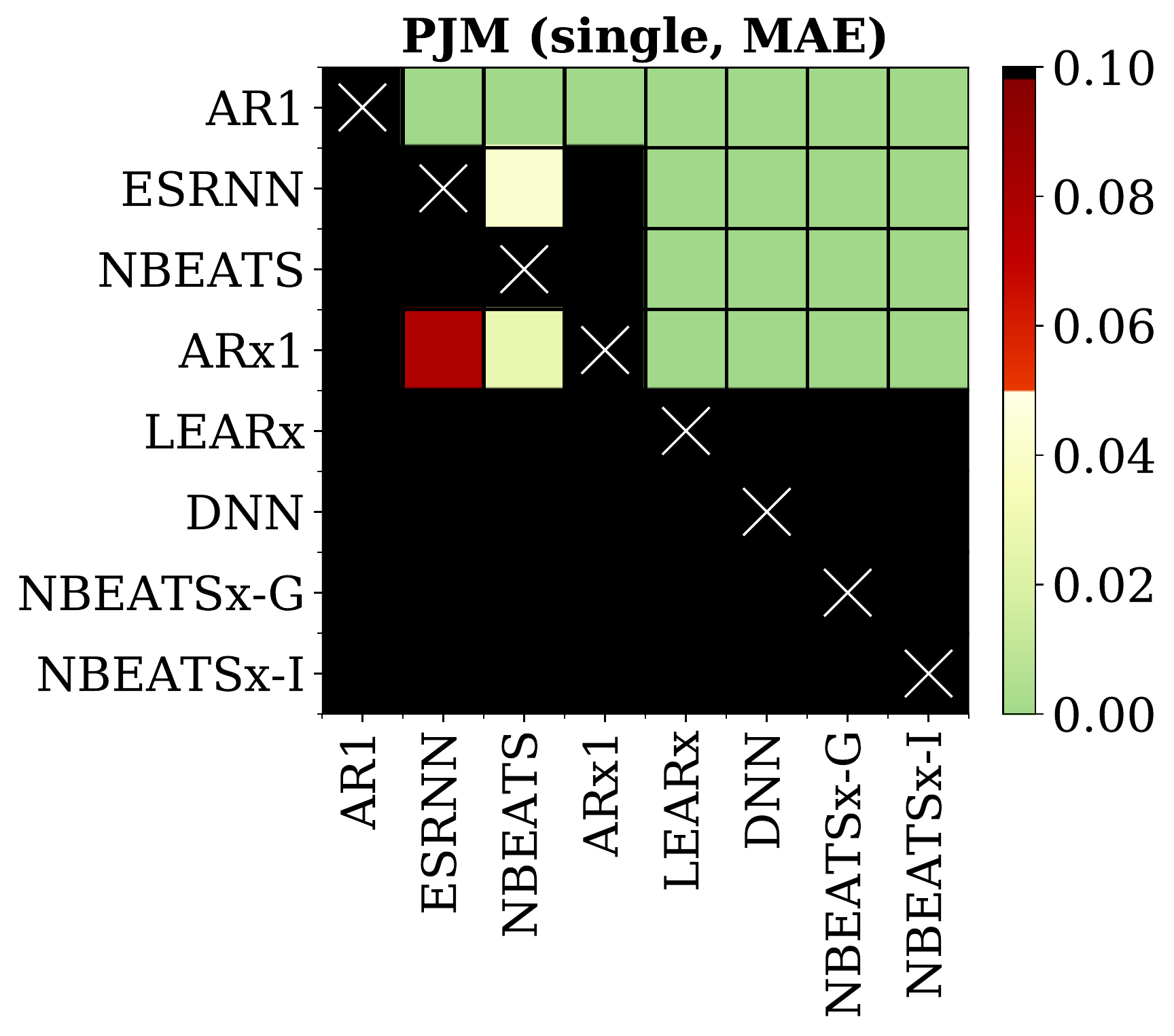}\quad
\includegraphics[width=.28\textwidth]{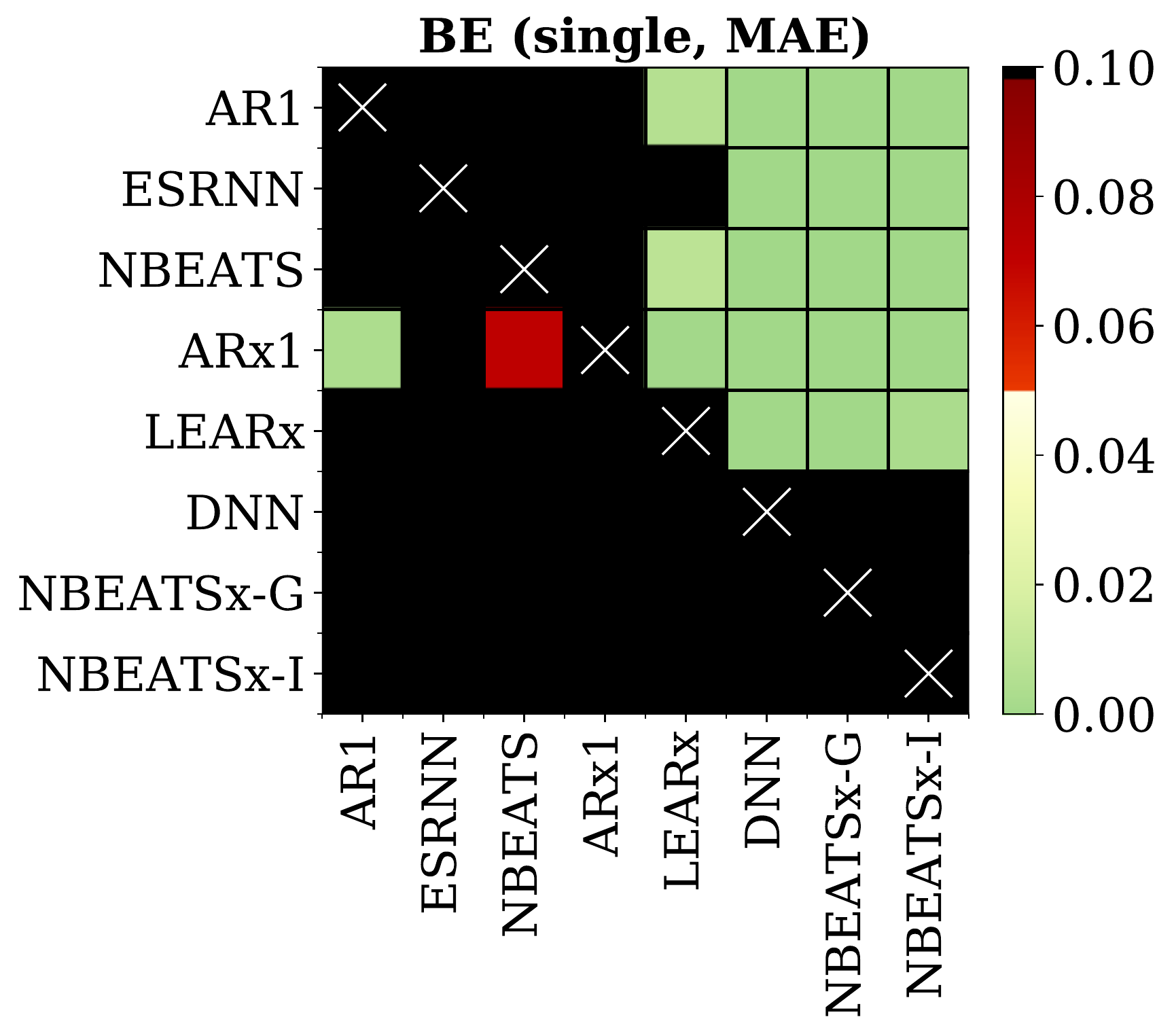}
\medskip
\includegraphics[width=.28\textwidth]{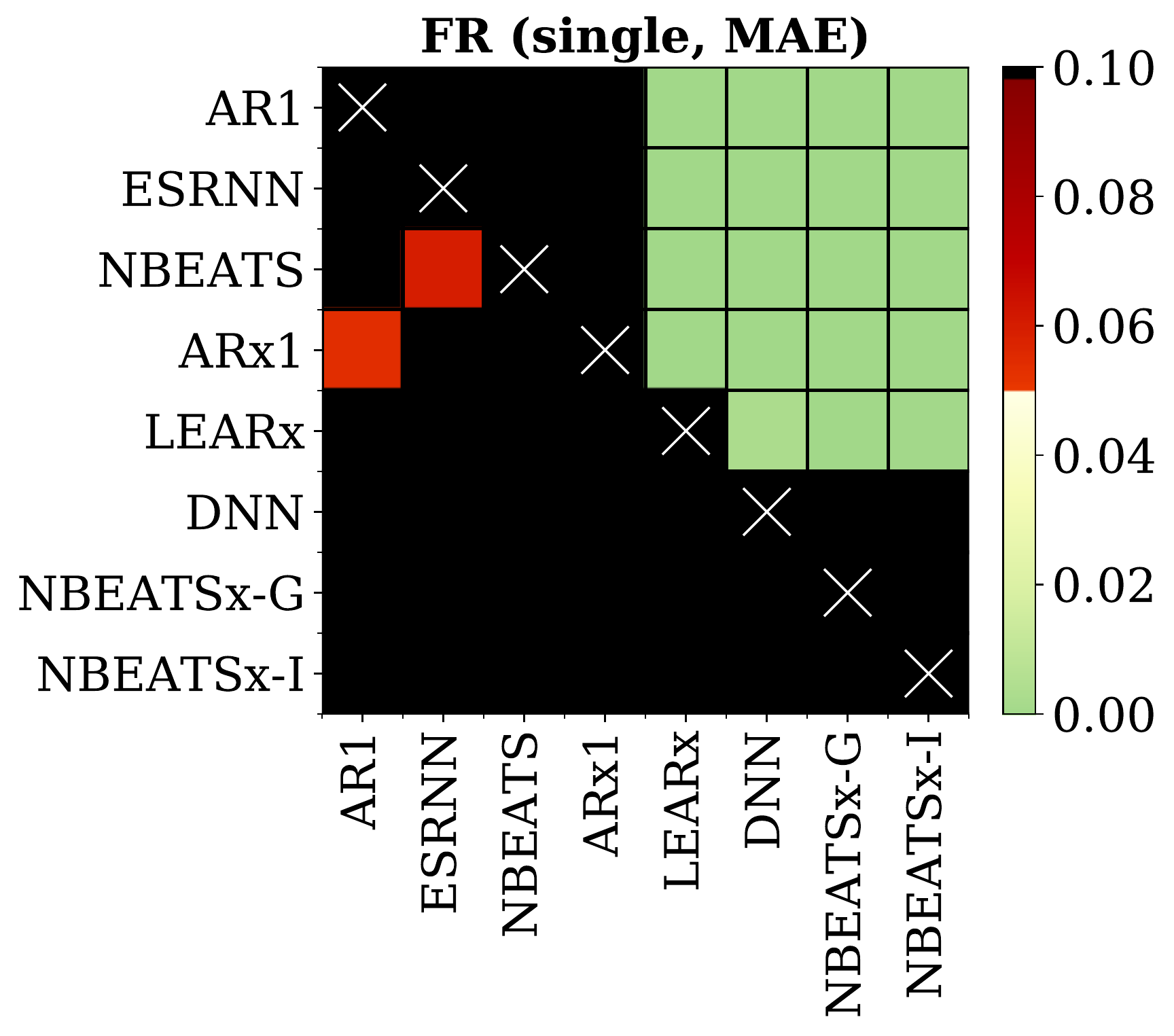}\quad
\includegraphics[width=.28\textwidth]{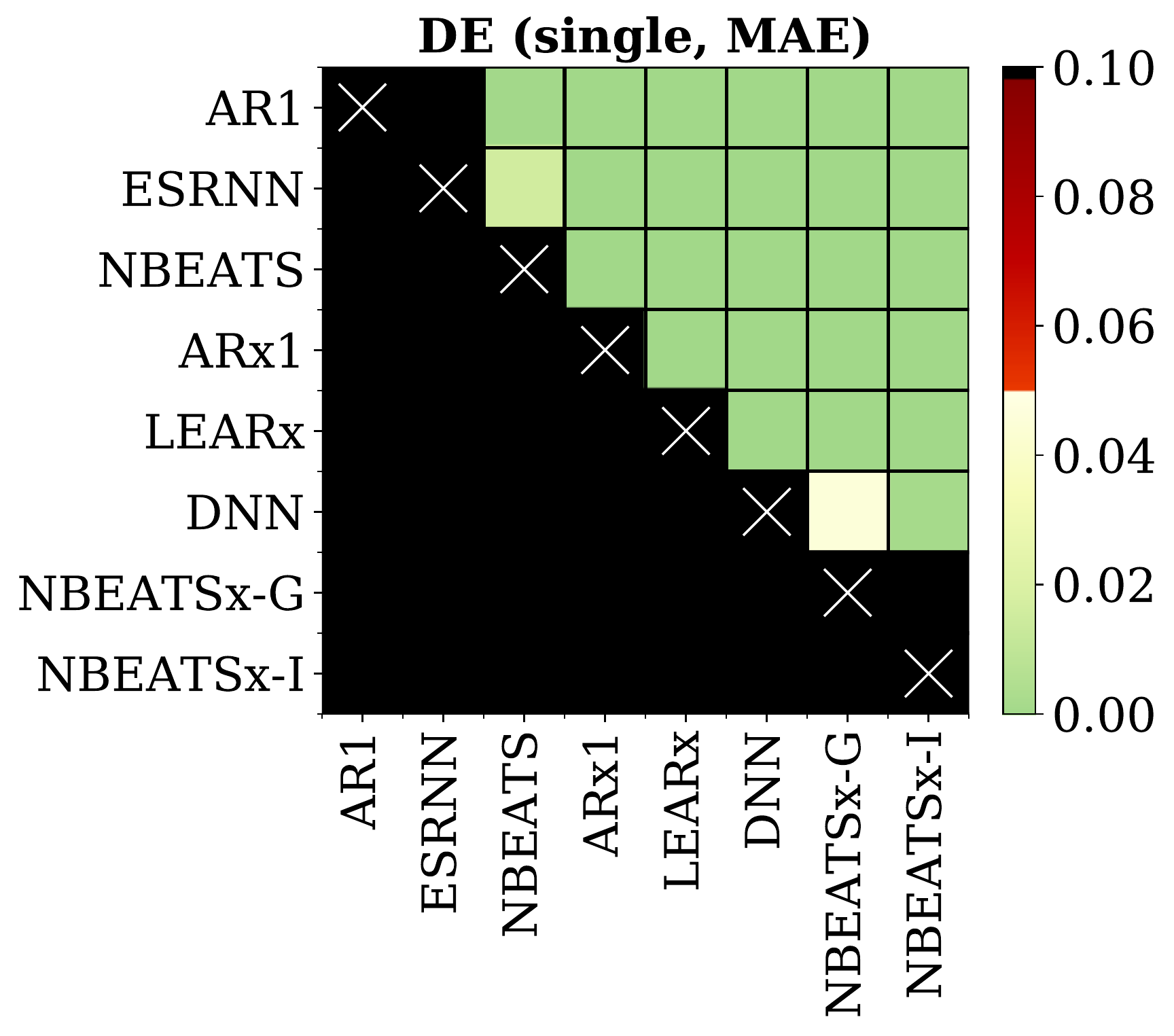}
\caption{Results of the Giacomini-White test for the day-ahead predictions with \emph{mean absolute error} (MAE) applied to pairs of the single models on the five electricity markets datasets. Each grid represents one market. Each colored cell in a grid is plotted black, unless the predictions of the model corresponding to its column of the grid outperforms the predictions of the model corresponding to its row of the grid. 
The color scale reflects significance of the difference in MAE, with solid green representing the lowest $p$-values.} \label{fig:pvals_singleMAE}
\end{figure}

Figure~\ref{fig:pvals_singleMAE} provides a graphical representation of the GW test for the six best models, across the five markets for the MAE evaluation metric. The models included in the significance tests are the same as in Tables~\ref{table:main_results_single}: \LEAR, \DNN, the \ESRNN, \NBEATS, and our proposed methods, the \NBEATSxg\ and \NBEATSxi. The p-value of each individual comparison shows if the improvement in performance (measured by MAE or RMSE) of the x-axis model over the y-axis model is statistically significant. Both the \NBEATSxg\ and \NBEATSxi\ model outperformed the \LEAR\ and \DNN\ models in all markets, with the exception of Belgium. Moreover, no benchmark model outperformed the \NBEATSxi\ and \NBEATSxg\ on any market.

\begin{table}[bp]
\caption{Forecast accuracy measures for day-ahead electricity prices for the best \emph{single model} out of the four models described in the Subsection~\ref{section:ensembling}. The \ESRNN\ and \NBEATS, are the original implementations and do not include time dependent covariates. The reported metrics are \emph{mean absolute error} (MAE), \emph{relative mean absolute error} (rMAE), \emph{symmetric mean absolute percentage error} (sMAPE) and \emph{root mean squared error} (RMSE). The smallest errors in each row are highlighted in bold.\\ \textsuperscript{*} The \LEARx \ results for \DE \ differ from \cite{lago2021epftoolbox} -- the values presented there are revised \citep{lago2021erratum}}.
\label{table:main_results_single}
\centering
\scriptsize
\begin{tabular}{llcccccccc}
                         &       & \ARone     & \ESRNN  & \NBEATS & \ARxone & \LEARx\textsuperscript{*} & \DNN            & \NBEATSxg           & \NBEATSxi      \\ \hline
\multirow{4}{*}{\NP}     & MAE   &   2.28     &  2.11   &  2.11 &   2.11    &   1.95                    & 1.71            & \textbf{1.65}       & 1.68           \\
                         & rMAE  &   0.72     &  0.67   &  0.67 &   0.67    &   0.62                    & 0.54            & \textbf{0.52}       & 0.53           \\
                         & sMAPE &   6.51     &  6.09   &  6.06 &   6.1     &   5.62                    & 4.97            & \textbf{4.83}       & 4.89           \\
                         & RMSE  &   4.08     &  3.92   &  3.98 &   3.84    &   3.60                    & 3.36            & \textbf{3.27}       & 3.33           \\ \hline
\multirow{4}{*}{\PJM}    & MAE   &   3.88     &  3.63   &  3.48 &   3.68    &   3.09                    & 3.07            & 3.02                & \textbf{3.01}  \\
                         & rMAE  &   0.8      & 0.75    &  0.72 &   0.76    &   0.64                    & 0.63            & \textbf{0.62}       & \textbf{0.62}  \\
                         & sMAPE &   14.66    & 14.26   & 13.56 &   14.09   &  12.54                    & 12.00           & 11.97               & \textbf{11.91} \\
                         & RMSE  &   6.26     &  5.87   &  5.59 &   5.94    &   5.14                    &  5.20           & 5.06                & \textbf{5.00}  \\ \hline
\multirow{4}{*}{\BE}     & MAE   &   7.04     &  7.01   &  6.83 &   7.05    &   6.59                    & \textbf{6.07}   & 6.14                & 6.17           \\
                         & rMAE  &   0.86     &  0.86   &  0.83 &   0.86    &  0.80                     & \textbf{0.74}   & 0.75                & 0.75           \\
                         & sMAPE &   16.29    &  15.95  & 16.03 &   16.21   &  15.95                    & \textbf{14.11}  & 14.68               & 14.52          \\
                         & RMSE  &   17.25    &  16.76  & 16.99 &   17.07   &  16.29                    & 15.95           & 15.46               & \textbf{15.43} \\ \hline
\multirow{4}{*}{\FR}     & MAE   &   4.74     &   4.68  &  4.79 &   4.85    &   4.25                    &  4.06           & 3.98                & \textbf{3.97}  \\
                         & rMAE  &   0.80     &   0.78  &  0.80 &   0.86    &  0.71                     & 0.68            & \textbf{0.67}       & \textbf{0.67}  \\
                         & sMAPE &   13.49    &  13.25  & 13.62 &   16.21   &  13.25                    & 11.49           & \textbf{11.07}      & 11.29          \\
                         & RMSE  &   13.68    & 11.89   & 12.09 &   17.07   &  \textbf{10.75}           & 11.77           & 11.61               & 11.08          \\ \hline
\multirow{4}{*}{\DE}     & MAE   &   5.73     & 5.64    & 5.37  &   4.58    &   3.93                    & 3.59            & 3.46                & \textbf{3.37}  \\
                         & rMAE  &   0.71     & 0.70    & 0.67  &   0.57    &  0.49                     & 0.45            & 0.43                & \textbf{0.42}  \\
                         & sMAPE &   21.22    & 21.09   & 19.71 &   18.52   &  16.80                    & 14.68           & 14.78               & \textbf{14.34} \\
                         & RMSE  &   9.39     & 9.17    & 9.03  &   7.69    &   6.53                    & 6.08            & 5.84                & \textbf{5.64}  \\ \hline
\end{tabular}
\end{table}

\subsection{Comments on Hyperparameter Optimization}

\EDITtwo{
In this Section, we summarize observations and key empirical findings from the extensive hyperparameter optimization on the space defined by Table~\ref{table:hyperparameters} for the four models composing each dataset ensemble. These observations and regularities of the optimally selected hyperparameters are important to create a more efficient and informed hyperparameter space and possibly guide future experiments with the \NBEATSx\  architecture.

\vspace{3mm}
\noindent
Interpretable configuration observations:
\begin{enumerate}
    \item Among quadratic, cubic and fourth degree polynomials, $N_{pol}\in \{2,3,4\}$, the most common  basis selected for the day-ahead EPF task was quadratic, $N_{pol}=2$. As shown in Figure~\ref{fig:forecast_decomposition}, the combination of quadratic trend and harmonics already describes the electricity price average daily profiles successfully. Linear trends were omitted from exploration as they showed to be fairly restrictive. In experiments on longer forecast horizons ($H>24$), beyond the scope of this paper, we observed that more trend flexibility tended to be beneficial.
    \item We did not observe preferences in the harmonic basis spectrum controlled by $N_{hr} \in \{1, 2\}$, the hyperparameter that controls the number of oscillations of the basis in the forecast horizon. We believe this is due to the flexibility of the harmonic basis $S \in \mathbb{R}^{H \times (H-1)}$ that already covers a broad spectrum of frequencies. Our intuition dictates that $N_{hr}=1$ is a good setting unless there is an apparent mismatch between the time-series frequency and the number of recorded observations that one could have in a Nyquist-frequency under-sampling or over-sampling phenomenon~\citep{koopmans1995spectral_timeseries}. This, however, is beyond the scope of this paper.
\end{enumerate}

\vspace{3mm}
\noindent
Hyperparameter optimization regularities:
\begin{enumerate}
    \item Regarding the optimal activation functions, we found that the most selected ones were \SeLU, \PreLU, and \Sigmoid, while activations like \ReLU, \TanH, and \LReLU\ were consistently outperformed. \Sigmoid\ activations tend to make the optimization of the network difficult when the networks grow in depth.
    \item Surprisingly, the stochastic gradient batch size consistently preferred 256 and 512 over 128 windows. Our selection of the \ADAM\ optimizer over classic \SGD\ could explain these observations. The machine learning community believes that more extended \SGD\ optimization with mini batches tends to have better generalization properties~\citep{keskar2017minibatch_generalization}. Additional research on the area would be interesting.
    \item The batch normalization technique was often detrimental in combination with the doubly-residual stack strategy of the \NBEATSx\ method. The residual signals tend to be close to zero, making the normalization numerically unstable.
    \item The robust median normalization of the exogenous variables was consistently preferred over alternatives like standard deviation normalization.
    \item Regarding the hidden units of the \FCNN\ layers, the optimal parameters did not favor an information bottleneck behavior \citep{tishby2000information_bottleneck}. Almost half of the optimal models had a small number of hidden units followed by a larger number of hidden units.
\end{enumerate}
}

\end{document}